\documentclass{article}

    \PassOptionsToPackage{numbers, compress, sort}{natbib}
 \usepackage[main, final]{neurips_2026}
\usepackage[utf8]{inputenc} % allow utf-8 input
\usepackage[T1]{fontenc}    % use 8-bit T1 fonts
\usepackage{hyperref}       % hyperlinks
\usepackage{url}            % simple URL typesetting
\usepackage{booktabs}       % professional-quality tables
\usepackage{amsfonts}       % blackboard math symbols
\usepackage{nicefrac}       % compact symbols for 1/2, etc.
\usepackage{microtype}      % microtypography
\usepackage{microtype}      % microtypography
\usepackage{xcolor}         % colors
\usepackage{graphicx}
\usepackage{subfigure}
\usepackage{setspace}
\usepackage{multicol}
\usepackage{multirow}
\usepackage{amsmath}
\usepackage{amssymb}
\usepackage{bbm}
\usepackage{hyperref}
\usepackage{color}
\usepackage{bbding}
\usepackage{wrapfig}
\usepackage{enumitem}
\usepackage{pifont}

\title{SteerVTE: Seamless Video Text Editing with Style and Glyph Control}

\vspace{-8mm}
\author{
\hspace{-3mm}
\textbf{Kai Zeng}\textsuperscript{\rm 1,2}\hspace{0.2cm}
\textbf{Moran Li}\textsuperscript{\rm 2}\hspace{0.2cm}
\textbf{Zhengwei Wang}\textsuperscript{\rm 2}\hspace{0.2cm}
\textbf{Yingchen Yu}\textsuperscript{\rm 3,$\S$}\hspace{0.2cm}
\textbf{Yiheng Lin}\textsuperscript{\rm 2}
\\[0.2cm]
\textbf{Ruichuan An}\textsuperscript{\rm 1}\hspace{0.2cm}
\textbf{Ming Lu}\textsuperscript{\rm 1}\hspace{0.2cm}
\textbf{Qi She}\textsuperscript{\rm 2,$\dagger$}\hspace{0.2cm}
\textbf{Wentao Zhang}\textsuperscript{\rm 1,4,5$\S$}
\\[0.3cm]
\textsuperscript{\rm 1} Peking University 
\textsuperscript{\rm 2} ByteDance China 
\textsuperscript{\rm 3} ByteDance Singapore
\textsuperscript{\rm 4} Zhongguancun Academy
\\[0.1cm]
\textsuperscript{\rm 5} Beijing Key Laboratory of Data Intelligence and Security (Peking University)
}
\vspace{-8mm}

\usepackage[table]{xcolor}
\usepackage{wrapfig}
\usepackage{booktabs}
\usepackage{cleveref}
\usepackage{makecell}

\newcommand{\Vsrc}{\mathbf{V}_{\mathrm{src}}}  % 源视频
\newcommand{\DiT}{v_\theta}
\newcommand{\VAEenc}{\mathcal{E}}              % VAE 编码器
\newcommand{\VAEdec}{\mathcal{D}}
\newcommand{\Text}{\mathbf{c}} % text condition
\newcommand{\Cond}{\mathbf{C}}                % 空间条件张量
\begin{document}

\maketitle

\begingroup
\renewcommand{\thefootnote}{\ensuremath{\dagger}}
\footnotetext{Project Leader.}
\renewcommand{\thefootnote}{\S}
\footnotetext{Corresponding Authors.
\texttt{yingchen001@e.ntu.edu.sg} and
\texttt{wentao.zhang@pku.edu.cn}.}
\endgroup

% \vspace{-9mm}
% \begin{figure}[h] % h表示here，即尽量放在当前位置
%     \centering
%     \includegraphics[width=1.0\textwidth]{figs/teaser_3.pdf} % 图片路径和大小
%     \vspace{-5mm}
%     \caption{STEER achieves precise video text editing while preserving style consistency. In contrast, previous SOTA video editing models lack editing region control and typically produce garbled text.}
%     \vspace{-2mm}
%     \label{fig:teaser}
% \end{figure}
\vspace{-4mm}
\begin{figure}[h] % h表示here，即尽量放在当前位置
    \centering
    \includegraphics[width=1.0\textwidth]{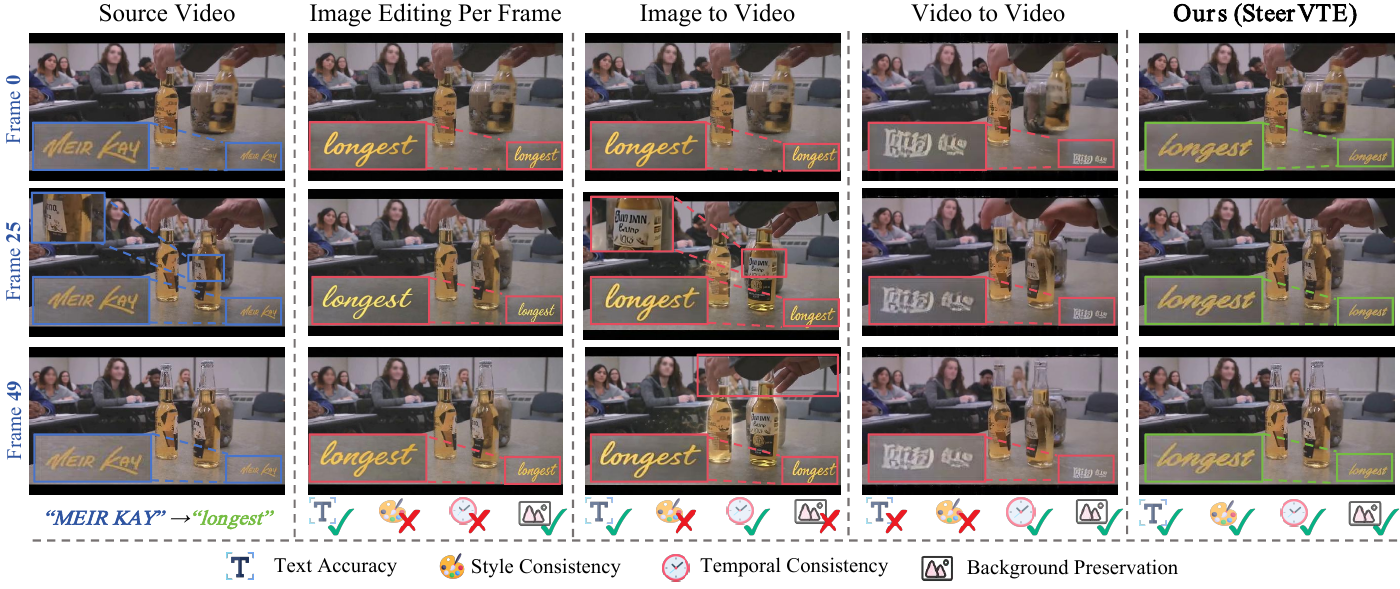} % 图片路径和大小
    % \vspace{-3mm}
    \caption{\textbf{Challenges in video text editing.} Existing approaches suffer from fundamental trade-offs: frame-wise image editing ~\cite{lan2025flux,tuo2024anytext2,wu2025qwen} ensures text accuracy but breaks style and temporal consistency. Image-to-video methods~\cite{kong2024hunyuanvideo,tongyi_wan27} improve temporal coherence but sacrifice content fidelity. General video editing ~\cite{cong2025viva,wei2025univideo,lin2026kiwi} lacks text priors and often produces garbled text. Our method tackles these trade-offs, enabling accurate, temporally consistent, and style-preserving video text editing.}
    % \vspace{-2mm}
    \label{fig:motivation}
\end{figure}

\begin{abstract}
Visual text editing aims to precisely modify text in images and videos while preserving stylistic consistency and visual realism. 
Despite significant advances in the image domain, video text editing remains largely unexplored: it is a localized task demanding stroke-level precision within small text regions, which compounds the challenges of cross-frame accuracy, temporal coherence, and stylistic fidelity. 
We introduce SteerVTE, a unified framework that \underline{\textbf{steer}}s a frozen video diffusion model to perform precise \underline{\textbf{V}}ideo \underline{\textbf{T}}ext \underline{\textbf{E}}diting through style and glyph control. 
Built on a frozen diffusion transformer, SteerVTE attaches a lightweight text context adapter with two complementary modules: a style encoder capturing the original text's visual attributes, and dual-granularity glyph encoders encoding the target text at both the line and character levels.
To overcome the inherently weak text rendering priors of video foundation models, we further propose a glyph-aware spatial-focal loss and a three-stage progressive training curriculum that scales from image to video data.
To support large-scale training, we also develop an automatic synthesis pipeline and construct SteerVTE-1M, a dataset of one million triplets spanning diverse scenes, fonts, and stylistic effects.
Extensive experiments demonstrate that SteerVTE substantially outperforms existing video editing baselines across text accuracy, style consistency, and temporal coherence.
Our project page is available at \url{https://zengkaiya.github.io/SteerVTE/}.
% \href{https://steervte-paper.github.io/}{https://steervte-paper.github.io}.
% Our model, dataset, and benchmark will be publicly released.
\end{abstract}

\section{Introduction}
% Visual text editing aims to modify text in natural images or videos while ensuring accurate and seamless background integration.
% This task has broad applications, such as advertisement design, in-game UI localization, film post-production, and short-video subtitle correction, benefiting both professional designers and everyday users.
% Accomplishing this task requires a model to understand text attributes (e.g., font, size, color) and to render glyphs accurately across diverse visual contexts.
% In the image domain, advances in diffusion models~\cite{ho2020denoising, song2020denoising, dhariwal2021diffusion}, powerful foundations such as Stable Diffusion 3~\cite{esser2024scaling}, FLUX.1~\cite{labs2025flux}, and Qwen-Image~\cite{wu2025qwen} have enabled strong text editing performance with minimal task-specific fine-tuning.

% Visual text editing aims to modify text in images or videos while ensuring accurate and seamless background integration. This task has broad applications including advertisement design, film post-production, and subtitle correction, benefiting both professionals and novice users. It requires models to understand text attributes such as font, size, and color, and to render glyphs accurately across diverse contexts. 
Visual text editing aims to modify text in images or videos while ensuring accurate and seamless background integration. This task has broad applications, including advertisement design, film post-production, and subtitle correction.
It requires models to understand text attributes such as font, size, and color, and to render glyphs accurately.
In the image domain, advances in diffusion models~\cite{ho2020denoising, song2020denoising, dhariwal2021diffusion} and powerful foundations like Stable Diffusion 3~\cite{esser2024scaling}, FLUX.1~\cite{labs2025flux}, and Qwen-Image~\cite{wu2025qwen} have enabled strong text editing performance with minimal task-specific fine-tuning.

However, in the video domain, seamless text editing demands not only accurate modifications and style consistency but also temporal coherence across frames. 
As illustrated in Fig.~\ref{fig:motivation}, straightforward solutions all fall short. Applying image text editing frame-by-frame ignores cross-frame dependencies, inducing temporal flickering and style drift.
Editing only the first frame and propagating it via image-to-video (I2V) generation enforces temporal consistency, but without conditioning on subsequent source frames, the model hallucinates motion and scene dynamics, breaking fidelity to the original.
End-to-end video-to-video (V2V) approaches appear promising, yet pre-trained video foundation models (e.g., LTXVideo~\cite{hacohen2024ltx}, Wan~\cite{wan2025wan}, HunyuanVideo~\cite{kong2024hunyuanvideo}) are optimized primarily for general scene generation rather than fine-grained text rendering. Consequently, V2V models built on these backbones tend to struggle with the character-level precision that text editing requires. Meanwhile, existing controllable video editing methods~\cite{cong2025viva, wei2025univideo, bian2025videopainter, jiang2025vace, lin2026kiwi} are tailored for global or object-level edits such as stylization and object insertion/deletion, and consequently offer limited support for capturing small-region text styles or generating stroke-level glyphs.

% To address these challenges, we propose \textbf{SteerVTE}, a unified framework that jointly tackles model architecture, training strategy, and large-scale data synthesis for video text editing. 
% SteerVTE builds upon a pre-trained text-to-video diffusion transformer (DiT) and keeps its backbone frozen to preserve the generative prior, while introducing a lightweight Text Context Adapter that supplies three complementary editing signals: (1) a spatiotemporal mask for precise region localization; (2) a style encoder, built on a native-resolution Vision-Language Model, for extracting clean stylistic features from the cropped editing region; (3) a dual-granularity glyph module fusing sentence-level and character-level representations for accurate text rendering.
% Together, these signals equip the frozen backbone with the region awareness, style fidelity, and content accuracy, enabling seamless video text editing across spatial, temporal, and semantic dimensions.

To break this impasse, we introduce \textbf{SteerVTE}, a framework that fundamentally rethinks video text editing along three axes: model architecture, training strategy, and large-scale data synthesis. 
At its core is a simple yet pivotal insight: rather than retraining a video diffusion backbone and risking the collapse of its rich generative prior, we freeze a pre-trained text-to-video DiT and, through a lightweight Text Context Adapter, infuse it with exactly the missing editing awareness. This adapter delivers three carefully designed signals: (1) a spatiotemporal mask for precise region localization; (2) a style encoder, built on a native-resolution Vision-Language Model, for extracting clean stylistic features from the cropped editing region; (3) a dual-granularity glyph module fusing sentence-level and character-level representations for accurate text rendering.
Together, these signals equip the frozen backbone with the region awareness, style fidelity, and content accuracy, enabling seamless video text editing across spatial, temporal, and semantic dimensions.
% ——a precise spatiotemporal mask for localized editing, a native-resolution VLM-based style encoder that extracts pristine stylistic detail from the target region, and a dual-granularity glyph module that fuses sentence- and character-level representations, enabling the backbone to render text not just correctly, but faithfully across space, time, and visual style.

Existing image text editors still struggle when scaling from single-word to sentence-level edits and small-font scenarios. We address this limitation with an \textbf{GL}yph-\textbf{A}ware \textbf{S}patial-focal Loss (\textbf{GLAS Loss}) that concentrates supervision on text regions and provides character-level guidance.
To overcome the inherently weak text rendering in video foundation models, we introduce a progressive \textbf{three-stage curriculum} that bridges image and video domains: the first stage establishes basic glyph priors on simple image text editing data; the second stage introduces complex real-world image data to learn font, color, and layout variations; the final stage fine-tunes on video data with the flow matching loss and GLAS loss for temporally stable editing. This curriculum learning strategy progressively builds robust text editing capabilities and avoids the difficulty of simultaneously mastering glyph rendering, stylistic fidelity, and temporal coherence directly on video data.

The three-stage curriculum above requires a corresponding data strategy, yet large-scale real video text editing pairs are unavailable. 
We therefore design a fully automatic synthesis pipeline with four stages: text configuration sampling, paired subtitle texture rendering (same style, different text), random video compositing, and OCR-based filtering to ensure clear and recognizable text regions. This pipeline involves no generative models, guaranteeing pixel-perfect editing accuracy and exact background consistency. However, models trained solely on synthetic data generalize poorly to real footage. 
To bridge this gap, we incorporate real-world image editing data derived from a raw OCR dataset, which provides stylistic diversity not present in synthetic renderings.
Together, the two sources form \textbf{SteerVTE-1M}, a one-million-triplet dataset that gives the curriculum both precise supervision from synthetic data and realistic appearance variations from real scenes.

% Beyond training data, evaluation is equally lacking: no dedicated benchmark exists for video text editing. We therefore construct \textbf{VTE-Bench}, the first such benchmark, comprising 100 synthetic and 100 real-world videos that systematically evaluate text accuracy, style consistency, temporal coherence, and background preservation. Quantitative and qualitative results demonstrate that SteerVTE achieves state-of-the-art performance, outperforming all open-source baselines and achieving over \textbf{2× higher} Sentence Accuracy than the proprietary model Seedance 2.0 (77\% vs. 32\%), along with a 34-point improvement on NED (95\% vs. 61\%) and a 25-point Style-Sim improvement (59\% vs. 34\%).

Beyond training data, evaluation is equally lacking: no dedicated benchmark exists for video text editing. We therefore construct \textbf{VTE-Bench}, the first benchmark of its kind, comprising 100 synthetic videos, 100 real-world videos, and 200 challenging scene text videos. VTE-Bench systematically evaluates text accuracy, style consistency, temporal coherence, and background preservation. Quantitative and qualitative results show that SteerVTE achieves state-of-the-art performance, outperforming all baselines. In particular, it achieves more than \textbf{twice} the Sentence Accuracy of the proprietary model Seedance 2.0 (77\% vs. 32\%), along with a 34-point improvement in NED (95\% vs. 61\%) and a 25-point improvement in Style-Sim (59\% vs. 34\%).

We summarize our contributions as follows:
\begin{itemize}[leftmargin=1em]
    \item We present SteerVTE, to our knowledge, the first \textbf{end-to-end} model for seamless video text editing. Built on a frozen DiT with a Text Context Adapter, it integrates a VLM-based style encoder and a dual-granularity glyph module to deliver style-preserving and content-accurate edits.
    % \item We propose GLAS Loss and a three-stage progressive training curriculum, which jointly improve performance on sentence-level and small-region text editing.
    \item We propose GLAS Loss and a three-stage curriculum to counteract the deficient text priors in video foundation models, jointly improving performance on sentence-level and small-region text editing.
    \item To address the lack of training data and standardized evaluation, we construct SteerVTE-1M, a one-million-triplet dataset combining synthetic and real-world supervision, and VTE-Bench, the first benchmark for video text editing spanning diverse scenarios.
\end{itemize}
\section{Related Work}
\textbf{Visual Generation Foundation Models.}
Diffusion models~\cite{ho2020denoising, song2020denoising, dhariwal2021diffusion, rombach2022high} have driven significant progress in image and video generation. 
% Early text-to-image models~\cite{nichol2021glide, ho2022imagen, ramesh2022hierarchical, podell2023sdxl} and text-to-video models~\cite{singer2022make, ho2022video, guo2023animatediff, blattmann2023stable} leveraged latent diffusion models (LDMs)~\cite{rombach2022high}. 
More recently, scalable diffusion transformers~\cite{peebles2023scalable} and flow-matching~\cite{esser2024scaling} have further improved generation quality and efficiency, enabling large-scale image foundation models (e.g., Flux~\cite{labs2025flux}, Qwen-Image~\cite{wu2025qwen}) and video foundation models (e.g., HunyuanVideo~\cite{kong2024hunyuanvideo}, Wan~\cite{wan2025wan}), which serve as backbones for a wide range of downstream tasks.

\textbf{Image Text Generation and Editing.}
The practical value of glyphs in applications like advertisements and films has driven growing interest in visual text generation and editing. Most existing works~\cite{chen2023textdiffuser, tuo2023anytext, zhu2024visual, chen2024textdiffuser, tuo2024anytext2, liu2024glyph, shi2025fonts} follow pathways similar to those for natural images, typically fine-tuning diffusion models~\cite{rombach2022high} in a parameter-efficient manner~\cite{hu2022lora}. Fine-grained text variations, including color, font, and layout, have inspired diverse research focuses. For instance, DiffUTE~\cite{chen2023diffute} and GlyphDraw~\cite{ma2023glyphdraw} embed glyph images into text embeddings, while Anytext2~\cite{tuo2024anytext2} uses multiple encoders to capture fine-grained attributes. More recently, FLUX-Text~\cite{lan2025flux}, UM-Text~\cite{ma2026text}, and UTDesign~\cite{zhao2025utdesign} adopt DiT architectures for enhanced text generation. Nevertheless, current image text editing models remain limited to single-word editing, and video text editing remains underexplored.

\textbf{Controllable Video Editing.}
Video editing tasks have evolved alongside advances in diffusion models. Early training-free methods~\cite{qi2023fatezero,geyer2023tokenflow,li2024vidtome,kara2024rave,cong2023flatten,ku2024anyv2v,fan2024videoshop} perform DDIM inversion and denoising without training but suffer from limited generalization, high latency, and reliance on paired captions. One-shot methods~\cite{wu2023tuneavideo,gu2024videoswap,ouyang2024codef,ouyang2024i2vedit,an2026concept,an2024mc} overfit to individual videos for better quality yet require costly per-video optimization. With text-to-video (T2V) and I2V foundation models~\cite{yang2024cogvideox, kong2024hunyuanvideo, wan2025wan}, recent works~\cite{brooks2023instructpix2pix,cheng2023consistentinsv2v,zhang2024effived,hu2024vivid,liu2024generativegenprop,zeng2026rethinking} build large-scale datasets for training. Leveraging MLLM-DiT architectures, these methods~\cite{wei2025univideo, tan2025omni, liao2025context, lin2026kiwi} achieve powerful multimodal reference generation under joint instruction and visual guidance. However, existing video generation models have extremely limited text rendering capabilities, let alone fine-grained editing that preserves content and style consistency. 
Another line of work, STRIVE~\cite{subramanian2021strive}, adopts a cascaded pipeline: it sequentially detects text regions of interest (ROIs), selects a reference frame, applies an image-based text replacement model, and propagates the edited reference result to other frames through geometry modeling. Its text editing capability therefore comes from an external image editor, and it is prone to cascading errors.
To address this gap, we propose the first end-to-end video text editing model.

% Our model provides detailed guidance for the editing process through a style encoder and a dual-granularity glyph encoder. 
% Additionally, the proposed OGR-Loss and curriculum learning strategy progressively build the model's text rendering and editing capabilities, thereby filling the gap in visual text editing within the video domain.
\section{Methodology}

\begin{figure}[h] % h表示here,即尽量放在当前位置
    \centering
    \includegraphics[width=0.95\textwidth]{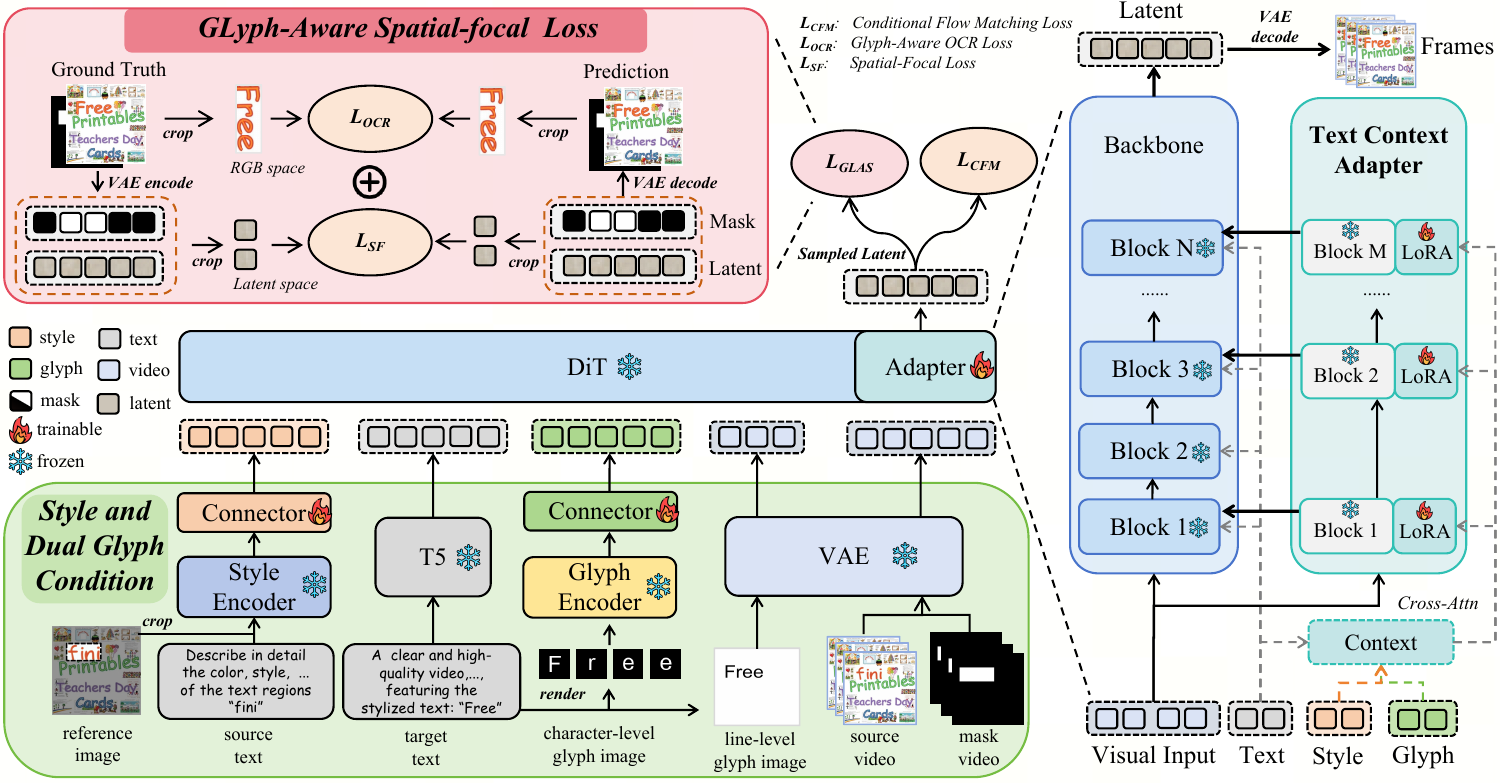} % 图片路径和大小
    % \vspace{-3mm}
    \caption{Overview of the proposed \textbf{SteerVTE}. A frozen text-to-video DiT is augmented with a lightweight Text Context Adapter that injects three editing signals via cross-attention: a spatiotemporal mask for region localization, VLM-encoded style features from a reference image cropped from the corresponding masked frame, and dual-granularity (line- and character-level) glyph features of the target text. The GLAS Loss further combines latent-space mask reweighting and RGB-space OCR supervision to improve sentence-level and small-region editing.}
    \vspace{-4mm}
    \label{fig:overview}
\end{figure}

The overview of SteerVTE is illustrated in Fig.~\ref{fig:overview}. We first detail the SteerVTE model architecture (Sec.~\ref{sec:model}) and its designated optimization objective, the GLyph-Aware Spatial-focal (GLAS) Loss (Sec.~\ref{sec:loss}). Subsequently, we introduce the supporting SteerVTE-1M dataset (Sec.~\ref{sec:dataset}) and the progressive Text Editing Curriculum Learning strategy tailored for stable optimization (Sec.~\ref{sec:training}).

\subsection{SteerVTE Model}
\label{sec:model}

\subsubsection{Preliminaries and Notation}
\textbf{Text-to-Video Generation Prior.} Text-to-video generation models are typically built upon Flow Matching (FM)~\cite{lipman2022flow} with a DiT backbone $\DiT$. 
To alleviate the computational overhead of pixel-space, these models operate in the latent space of a pretrained Variational Autoencoder (VAE) (i.e., encoder $\VAEenc$, decoder $\VAEdec$). 
Given a source video $\Vsrc$, Gaussian noise $\mathbf{z}_0 \sim \mathcal{N}(\mathbf{0}, \mathbf{I})$, and a timestep $t \in [0, 1]$, the noisy latent $\mathbf{z}_t$ is constructed via interpolation $\mathbf{z}_t = t \mathbf{z}_1 + (1-t) \mathbf{z}_0$, where $\mathbf{z}_1 = \VAEenc(\Vsrc)$. Conditioned on a text prompt $\Text$, $\DiT$ is optimized to estimate the velocity field driving the transition from noise to real latents. The general flow-matching objective is formulated as:
\begin{equation}
\mathcal{L}_{\text{FM}} = \| v_\theta(\mathbf{z}_t, t; \Text) - (\mathbf{z}_1 - \mathbf{z}_0) \|_2^2.
\end{equation}

\textbf{Context Adapter.} To leverage strong pretrained text-to-video priors for controllable generation, additional spatiotemporal conditions $\Cond$ are incorporated. Instead of employing computationally expensive parallel architectures like ControlNet~\cite{zhang2023adding}, we adopt a parameter-efficient approach utilizing Low-Rank Adaptation (LoRA)~\cite{hu2022lora}. Specifically, a lightweight adapter $\mathcal{A}_\phi$ extracts features from $\Cond$, which are subsequently injected into the DiT modules. 
The conditional flow-matching loss is:
\begin{equation}
\mathcal{L}_{\text{CFM}} = \| \DiT(\mathbf{z_t}, t; \Text, \mathcal{A}_\phi(\Cond)) - (\mathbf{z}_1 - \mathbf{z}_0) \|_2^2.
\end{equation}
% During training, only the adapter and the LoRA (collectively denoted by the trainable parameters $\phi$) are optimized.

\subsubsection{Text Context Adapter}
While general adapters suffice for coarse spatial control, they struggle in fine-grained video text editing, failing to precisely localize text, losing typographical attributes (e.g., font, color, and stroke), and yielding blurred or misspelled glyphs. To overcome this, we introduce the \textit{Text Context Adapter}, a dedicated conditioning module that jointly encodes positional, stylistic, and glyph-level cues (Fig.~\ref{fig:overview}).

\textbf{Mask-Guided Positional Conditioning.} To enable precise foreground editing while preserving background fidelity, following VACE~\cite{jiang2025vace}, we decompose the source video $\mathbf{V}_{\mathrm{src}}$ into foreground and background components via mask $\mathbf{M}$: $\mathbf{V}_{\mathrm{f}} = \mathbf{V}_{\mathrm{src}} \odot \mathbf{M}$, $\mathbf{V}_{\mathrm{b}} = \mathbf{V}_{\mathrm{src}} \odot (\mathbf{1} - \mathbf{M})$. Both components are encoded by the VAE encoder $\mathcal{E}$ and channel-wise concatenated with the down-sampled mask $\tilde{\mathbf{M}}$ for spatiotemporal alignment, yielding the positional condition: $\mathbf{C}_{\text{pos}} = \operatorname{Concat}(\mathcal{E}(\mathbf{V}_{\mathrm{b}}), \mathcal{E}(\mathbf{V}_{\mathrm{f}}), \tilde{\mathbf{M}})$.

\textbf{Native-Resolution VLM for Style Encoding.} Faithful style transfer relies on preserving the geometric and textural cues of source typography, such as stroke thickness, character spacing, and aspect ratio, which are highly sensitive to spatial distortion. 
% Prior text editors either learn style implicitly via masked inpainting or extract features with CLIP~\cite{radford2021learning}/DINO~\cite{oquab2023dinov2} encoders, both requiring square-resized inputs that warp the elongated geometry of text and collapse fine-grained typographical cues. 
Prior methods learn style implicitly via masked inpainting, lacking explicit control. To inject target style features, we introduce a dedicated style encoder. However, conventional encoders like CLIP~\cite{radford2021learning} or DINO~\cite{oquab2023dinov2} operate on square-resized inputs, warping the elongated geometry of text and collapsing fine-grained typographical cues. 
We argue that this limitation stems not from feature quality, but from fixed-resolution tokenization. Instead, we leverage native-resolution Vision-Language Models (VLMs) like Qwen2.5-VL~\cite{bai2025qwen3}, which tokenize images at their original aspect ratio and thus preserve the intrinsic spatial statistics of text, making them naturally suited for capturing typographic style across arbitrary shapes and orientations. 
Assuming stylistic temporal consistency within consecutive frames, we extract style from an arbitrary frame $k$: $\mathbf{C}_{\text{style}} = \text{VLM}(\mathbf{V}_{\mathrm{src}}^k \odot \mathbf{M}^k, P(\mathbf{T}_{\text{src}}))$, where $P(\cdot)$ is a prompt template steering the VLM toward the stylistic attributes of the source text $\mathbf{T}_{\text{src}}$. The final two VLM hidden states are concatenated and projected via a style connector for cross-attention into the $\DiT$.

\textbf{Dual-Granularity Glyph Control.}
Accurate text rendering hinges on three coupled questions: \textit{where} each character is placed, \textit{what} its strokes look like, and \textit{how} it is stylized. Stylization is handled by the VLM style encoding introduced previously, whereas structural placement and stroke fidelity remain underserved by the diffusion backbone. We therefore inject explicit glyph guidance at two complementary granularities, both rendered in a \emph{single fixed neutral font} so that the branches expose only content and structure, while font style is modeled exclusively by the VLM style encoder.
\emph{i). Line-level layout prior.} We render the target text $\mathbf{T}_{\text{tgt}}$ to obtain $\mathbf{I}_{\text{layout}} = \mathcal{R}_{\text{line}}(\mathbf{T}_{\text{tgt}})$, encode it via the VAE to $\mathbf{C}_{\text{layout}} = \mathcal{E}(\mathbf{I}_{\text{layout}})$, and concatenate it with the positional condition, $\mathbf{C} = \operatorname{Concat}(\mathbf{C}_{\text{pos}}, \mathbf{C}_{\text{layout}})$, yielding a spatially aligned, style-agnostic prior that dictates \emph{where} glyphs appear.
\emph{ii). Character-level stroke prior.}
Since VAE compression inevitably erodes fine strokes, the layout prior alone cannot prevent character-level distortions. We thus render each character $g_i \in \mathit{G}_{\text{tgt}} = \{g_1, \ldots, g_N\}$ individually. 
% To explicitly capture the canonical stroke topology, we encode them using a pre-trained OCR network $\mathcal{O}$~\cite{cui2025paddleocr} and inject the features via cross-attention: $\mathbf{C}_{\text{stroke}} = f_{\text{glyph}}\!\left( \operatorname{Concat}_{i=1}^{N} \mathcal{O}(\mathcal{R}_{\text{char}}(g_i)) \right)$. Sharing the neutral font ensures this prior provides precise \textit{what} guidance without leaking specific typeface details.
To explicitly capture the canonical stroke topology, we encode characters using a pre-trained OCR network $\mathcal{O}$~\cite{cui2025paddleocr} and pass through a glyph connector. The resulting stroke features are then injected via cross-attention: $\mathbf{C}_{\text{stroke}} = \text{Proj}_{\text{glyph}}\!\left( \operatorname{Concat}_{i=1}^{N} \mathcal{O}(\mathcal{R}_{\text{char}}(g_i)) \right)$. Sharing the neutral font ensures this prior provides precise \textit{what} guidance without leaking specific typeface details.
% \emph{iii). Style learning by design.}
% By construction, neither prior carries font-style information: stylistic variations in training images, e.g., serif/sans-serif, weight, slant, or decoration, cannot be explained by the glyph branches and must be attributed to the prompt and the diffusion prior. This decoupling of \emph{content} from \emph{style} forms a clean inductive bias, letting the two branches resolve the \textit{where} and \textit{what} of glyphs while style emerges as a factor learned from data.

Altogether, the Text Context Adapter aggregates position, style, and glyph signals into a unified conditioning representation. The base training objective is the conditional flow-matching loss $\mathcal{L}_{\text{CFM}}$.

% \subsection{Glyph-Aware Spatial-Focal (GLAS) Loss}
\subsection{GLyph-Aware Spatial-focal Loss}
\label{sec:loss}
Standard flow matching optimizes a training objective exclusively within the latent space, treating all spatial locations indiscriminately. 
This inherently introduces two critical bottlenecks for precise text editing: \emph{i). optimization bias}, in that gradients originating from spatially sparse textual regions are suppressed by the dominant non-text background; and \emph{ii). structural ambiguity}, i.e., global distribution matching in the latent space provides no explicit inductive bias toward fine-grained character topology; 
To overcome these limitations, we propose the \textbf{GL}yph-\textbf{A}ware \textbf{S}patial-focal (GLAS) loss (see Fig.~\ref{fig:overview}). GLAS instantiates a synergistic dual-space paradigm: a \textit{glyph-aware} OCR-based supervision in the pixel space that enforces character-level structural fidelity, coupled with a \textit{spatial-focal} reweighting in the latent space that prioritizes text-bearing regions.

\textbf{Spatial-Focal Loss.} To counteract spatial gradient suppression, we reweight the flow matching objective with a normalized focal map $\mathbf{W} \in [0,1]$ derived from the spatial mask $\mathbf{M}$, concentrating the optimization on text regions:
\begin{equation}
\mathcal{L}_{\text{SF}} = \bigl\| \mathbf{W} \odot \bigl(v_\theta(\cdot) - (\mathbf{z}_1 - \mathbf{z}_0)\bigr) \bigr\|_2^2.
\end{equation}

\textbf{Glyph-Aware OCR Loss.} The spatial-focal loss prioritizes text regions yet remains blind to glyph structure, since latent-space supervision cannot reach character-level topology. We thus impose pixel-space supervision via an off-the-shelf OCR recognizer $\mathcal{O}$. Specifically, the predicted latent $\hat{\mathbf{z}}_0$ is decoded and temporally sampled to obtain aligned frame triplets: $(\mathbf{I}_{\mathrm{gt}}, \mathbf{I}_{\mathrm{pred}}, \mathbf{I}_{\mathrm{m}}) = \mathcal{S}\bigl(\mathbf{V}_{\mathrm{gt}},\, \mathcal{D}(\hat{\mathbf{z}}_0),\, \mathbf{M}\bigr)$, where $\mathcal{S}(\cdot)$ uniform temporal sampling. Upon these triplets, we enforce glyph correctness at two complementary granularities:
\begin{equation}
\mathcal{L}_{\text{OCR}} = \underbrace{\bigl\| \mathcal{O}(\mathbf{I}_{\mathrm{pred}}) - \mathcal{O}(\mathbf{I}_{\mathrm{gt}}) \bigr\|_2^2}_{\mathcal{L}_{\text{text}}:\ \text{stroke-level fidelity}} \;+\; \underbrace{\mathrm{CTC}\bigl(\mathcal{O}(\mathbf{I}_{\mathrm{pred}}),\, y\bigr)}_{\mathcal{L}_{\text{CTC}}:\ \text{character-level recognizability}},
\end{equation}
where $y$ denotes the target transcription and $\mathrm{CTC}(\cdot)$ the connectionist temporal classification objective~\cite{graves2006connectionist}. The choice of recognizer and full formulations are deferred to Appx.~\ref{appx:loss_details}.

Hence, with empirical hyperparameters ($\gamma=3.0$, $\lambda=0.1$), the overall GLAS Loss is formulated as:
\begin{equation}
\mathcal{L}_{\text{GLAS}} = \gamma \mathcal{L}_{\text{SF}} +  \lambda \mathcal{L}_{\text{OCR}}.
\end{equation}

\begin{figure}[h] % h表示here,即尽量放在当前位置
    \centering
    \includegraphics[width=0.9\textwidth]{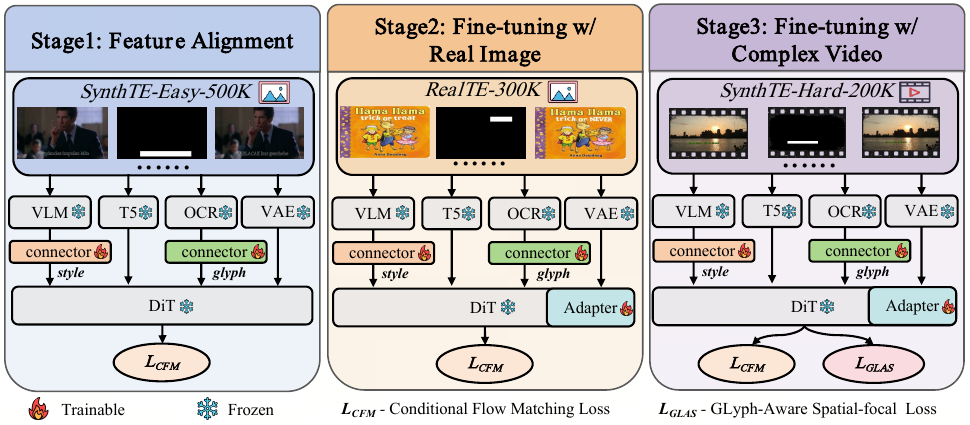} % 图片路径和大小
    % \vspace{-3mm}
    \caption{\textbf{Illustration of the three training stages.} In the first stage, the connector is primarily trained to align style and glyph features. In the second stage, the connector and DiT are jointly trained. In the third stage, the GLAS loss is introduced, and training proceeds on the video dataset.}
    \vspace{-5mm}
    \label{fig:training}
\end{figure}

\subsection{Dataset Construction}
\label{sec:dataset}
To empower end-to-end video text editing, we construct SteerVTE-1M, a large-scale dataset comprising two complementary components: SynthTE (stylized synthetic data) and RealTE (complex real-world data). Detailed construction pipelines are provided in Appx.~\ref{appx:dataset}.

\textbf{SynthTE Dataset.} To compensate for the weak text-rendering capabilities of video foundation models, we synthesize data with varying difficulty levels. We collect 1,933 TrueType fonts from Google Fonts~\cite{google_fonts}, an English text corpus~\cite{google_10000_english} of 65 letters and symbols, and background videos from OpenSora-Plan~\cite{lin2024open}. The easy version uses simple fonts with short text and large font sizes. The hard version uses all available fonts, including artistic and textured styles, featuring longer text and smaller font sizes. As illustrated in Fig.~\ref{fig:dataset_app}(a), the pipeline randomly samples text attributes and paired text to render subtitle images, selects background videos, and generates bounding boxes covering 8\% to 15\% of the frame for pasting. It inserts one to four subtitle pairs, selects one as the training region, and strictly retains only samples that achieve 100\% OCR recognition accuracy. This process produces 500k image pairs for SynthTE-Easy and 200k video pairs for SynthTE-Hard.

\textbf{RealTE Dataset.} Given the scarcity of real-world video text editing data, we construct an image-based dataset derived from AnyWord-3M~\cite{tuo2023anytext}. We filter 500k samples with fully legible text via PP-OCRv3, edit them using a state-of-the-art text editing model~\cite{zeng2024textctrl}, and reapply OCR verification. To preserve stylistic consistency, we employ a Text Style Similarity metric~\cite{seo2025stellar} and discard samples with scores below 0.4. 
This pipeline yields the RealTE Dataset, comprising 300k image pairs.
% This pipeline yields 300k high-quality image pairs.
\vspace{-3mm}
\subsection{Text Editing Curriculum Learning}
\label{sec:training}
Training video text editing from scratch is challenging. We therefore propose a progressive three-stage curriculum, leveraging our SteerVTE-1M dataset, to systematically transition from fundamental glyph alignment to complex temporal dynamics.

\textbf{Stage 1: Feature Alignment.} To establish an initial visual-textual mapping, we freeze the style and glyph encoders, training only their linear connectors. The Text Context Adapter is optimized via the standard flow-matching loss on 500k synthetic images (SynthTE-Easy).

\textbf{Stage 2: In-the-Wild Image Fine-Tuning.} To bridge the synthetic-to-real domain gap, we activate LoRA modules within the adapter. Both the connectors and LoRAs are jointly optimized on 300k real-world samples (RealTE), adapting the model to complex typographical layouts that synthetic data inherently lacks. 
Although the ground truth in this subset is generated by an image text editing model, rigorous OCR and style filtering ensure sample quality. We validate the effectiveness in Tab.~\ref{tab:ablation_training}.

\textbf{Stage 3: Temporal Video Fine-Tuning.} Image-only pretraining inevitably introduces temporal flickering. To enforce kinematic coherence, we extend training to 200k video clips (SynthTE-Hard). Having acquired robust foundational character priors from the previous stages, we seamlessly integrate the GLAS loss here to explicitly enforce spatial focus and fine-grained glyph fidelity. The ultimate optimization objective is: $\mathcal{L} = \mathcal{L}_{\text{CFM}} + \mathcal{L}_{\text{GLAS}}.$
\section{Experiments}
\label{sec:experiment}
% \subsection{ Experiment Settings}
\subsection{Experimental Setup}
\textbf{Benchmark and Metrics.} We construct \textbf{VTE-Bench}, a benchmark for video text editing, comprising VTE-Bench-Synth (50 easy and 50 hard synthetic samples with ground-truth edited videos), VTE-Bench-Real (100 real-world samples without ground truth), and VTE-Bench-SceneText (200 hard scene text videos from STRIVE~\cite{subramanian2021strive} and the web). Each sample provides a source video, 
editing mask, and editing prompt. We evaluate \textit{text 
accuracy} via PP-OCRv4 (precision, recall, F-score, NED, sentence 
accuracy) and \textit{video quality} via SSIM, PSNR, FID, Style-Sim, 
FVD, plus background preservation metrics SSIM$_{bg}$ and PSNR$_{bg}$. The evaluation OCR (PP-OCRv4) differs from the PP-OCRv3 used during training (Sec.~\ref{sec:loss}) to ensure independent measurement. A user study 
on VTE-Bench-Real provides an additional Quality Rate. More details are provided in Appx.~\ref{appx:benchmark}.

\textbf{Implementation.} 
In Stage 1, we initialize the style encoder, character glyph encoder, and backbone with Qwen2.5-VL-3B~\cite{bai2025qwen3}, PP-OCRv3~\cite{cui2025paddleocr}, and VACE 14B~\cite{jiang2025vace}, respectively, and train only the two randomly-initialized connectors. In Stages 2 and 3, we additionally fine-tune the Text Context Adapter via LoRA~\cite{hu2022lora} (rank 64). Full details in Appx.~\ref{appx:implementation}.

\subsection{Quantitative Evaluation}
\begin{table*}[h]
\centering
\scriptsize
\setlength{\tabcolsep}{3.2pt}
\renewcommand{\arraystretch}{1.10}
\vspace{-4mm}
\caption{\textbf{Quantitative comparison on VTE-Bench-Synth.} SteerVTE consistently outperforms other approaches across all metrics, including OCR accuracy and style consistency. \textbf{Bold} is the best.}
\vspace{2mm}
\resizebox{\textwidth}{!}{
\begin{tabular}{l|ccccc|ccccccc}
\toprule
% \multirow{2}{*}{Method}
\textbf{Method}
& \multicolumn{5}{c|}{\textbf{Text Editing}}
& \multicolumn{7}{c}{\textbf{Video Quality}} \\
\cmidrule(lr){2-6} \cmidrule(lr){7-13}
& Precision$\uparrow$
& Recall$\uparrow$
& F-Score$\uparrow$
& NED$\uparrow$
& Accuracy$\uparrow$
& SSIM$\uparrow$
& PSNR$\uparrow$
& FID$\downarrow$
& FVD$\downarrow$
& Style-Sim$\uparrow$
& SSIM$_{bg}$$\uparrow$
& PSNR$_{bg}$$\uparrow$ \\
\midrule
FLUX-Text~\cite{lan2025flux}
& 0.5601
& 0.5303
& 0.5447
& 0.2790
& -
& 0.2268
& 11.02
& 149.7
& 314.9
& 0.4273
& 0.9589
& 36.14 \\
VACE~\cite{jiang2025vace}
& 0.6168
& 0.6082
& 0.6124
& 0.1533
& -
& 0.2841
& 11.54
& 105.5
& 14.80
& 0.4827
& 0.9579
& 35.79 \\
VideoPainter~\cite{bian2025videopainter}
& 0.1653
& 0.1111
& 0.1328
& 0.0367
& -
& 0.1313
& 7.463
& 206.0
& 62.24
& 0.2446
& 0.3070
& 8.031 \\
VIVA~\cite{cong2025viva}
& 0.3446
& 0.2933
& 0.3168
& 0.1126
& -
& 0.3148
& 11.99
& 160.1
& 30.88
& 0.3570
& 0.9007
& 28.04 \\
UniVideo~\cite{wei2025univideo}
& 0.5480
& 0.5496
& 0.5487
& 0.1523
& -
& 0.2162
& 10.74
& 104.8
& 15.54
& 0.4476
& 0.6446
& 21.19 \\
Kiwi-Edit~\cite{lin2026kiwi}
& 0.5417
& 0.4839
& 0.5111
& 0.1280
& -
& 0.2078
& 10.58
& 118.7
& 18.81
& 0.4268
& 0.7265
& 21.85 \\
Seedance 2.0~\cite{seedance2026seedance}
& 0.9416
& 0.9144
& 0.9278
& 0.8628
& 0.3300
& 0.3203
& 12.10
& 121.1
& 54.71
& 0.3442
& 0.7015
& 22.30 \\
\rowcolor[RGB]{237,244,245}
SteerVTE (Ours)
& \textbf{0.9908}
& \textbf{0.9825}
& \textbf{0.9866}
& \textbf{0.9624}
& \textbf{0.6100}
& \textbf{0.4034}
& \textbf{12.30}
& \textbf{65.46}
& \textbf{14.55}
& \textbf{0.5936}
& \textbf{0.9606}
& \textbf{37.09} \\
\bottomrule
\end{tabular}
}
\label{tab:synbench}
\end{table*}
We compare SteerVTE with frame-by-frame image editors and video editors. Image editors include FLUX-Text~\cite{lan2025flux}; video editors include mask-based (VACE~\cite{jiang2025vace}, VideoPainter~\cite{bian2025videopainter}), instruction- and reference-guided (VIVA~\cite{cong2025viva}, UniVideo~\cite{wei2025univideo}, Kiwi-Edit~\cite{lin2026kiwi}), and closed-source (Seedance 2.0~\cite{seedance2026seedance}) models.
Results on VTE-Bench-Synth and VTE-Bench-Real are shown in Tab.~\ref{tab:synbench} and Tab.~\ref{tab:realbench}. For text accuracy, SteerVTE achieves near-perfect Precision, Recall, F-Score, and NED. On the stricter sentence-level accuracy, which requires exact character matching, SteerVTE reaches 61\% on Synth and 77\% on Real, more than doubling Seedance 2.0 with only 33\% and 32\%, while other video baselines lack text priors and perform poorly. For style and temporal coherence, SteerVTE leads on foreground style metrics and achieves an FVD of 14.55 on Synth, whereas FLUX-Text's per-frame editing causes severe flickering reflected by an FVD of 314.9. Our mask conditioning further enables near-perfect background preservation. 
Since Seedance 2.0 is the only baseline producing competitive editing results, we further conduct a user study to measure Quality Rate, defined as the fraction of videos judged visually satisfactory. SteerVTE achieves 72\% compared to only 23\% for Seedance 2.0. In addition, as shown in Tab.~\ref{tab:scenetextbench}, although our model is trained solely on real image data and synthetic video data, it generalizes to a wide range of static, dynamic, and deformable scene text cases.

\begin{table*}[h]
\centering
\scriptsize
\setlength{\tabcolsep}{4.2pt}
\renewcommand{\arraystretch}{1.10}
\caption{\textbf{Quantitative comparison on VTE-Bench-Real.} Other methods perform poorly under complex real-world scenarios, whereas SteerVTE achieves strong performance. \textbf{Bold} is the best.}
\vspace{2mm}
\resizebox{0.92\textwidth}{!}{
\begin{tabular}{l|ccccc|ccc}
\toprule
% \multirow{3}{*}{Method}
\textbf{Method}
& \multicolumn{5}{c|}{\textbf{Text Editing}}
& \multicolumn{3}{c}{\textbf{Video Quality}} \\
\cmidrule(lr){2-6} \cmidrule(lr){7-9}
& Precision$\uparrow$
& Recall$\uparrow$
& F-Score$\uparrow$
& NED$\uparrow$
& Accuracy$\uparrow$
& SSIM$_{bg}\uparrow$
& PSNR$_{bg}\uparrow$
& Quality Rate$\uparrow$ \\
\midrule
FLUX-Text~\cite{lan2025flux}
& 0.6221
& 0.6058
& 0.6138
& 0.3808
& -
& 0.9609
&33.40
& - \\
VACE~\cite{jiang2025vace}
& 0.4224
& 0.4185
& 0.4204
& 0.1082
& -
& 0.9626
& 34.33
& - \\
VideoPainter~\cite{bian2025videopainter}
& 0.0667
& 0.0385
& 0.0488
& 0.0113
& -
& 0.3044
& 7.965
& - \\
VIVA~\cite{cong2025viva}
& 0.4030
& 0.3704
& 0.3860
& 0.1680
& -
& 0.8688
& 22.26
& - \\
UniVideo~\cite{wei2025univideo}
& 0.3783
& 0.3910
& 0.3845
& 0.1197
& -
& 0.7399
& 20.66
& - \\
Kiwi-Edit~\cite{lin2026kiwi}
& 0.3892
& 0.3651
& 0.3767
& 0.0978
& -
& 0.7277
& 20.24
& - \\
Seedance 2.0~\cite{seedance2026seedance}
& 0.7577
& 0.7400
& 0.7487
& 0.6106
& 0.3200
& 0.7779
& 22.37
& 0.2300 \\
% \rowcolor{gray!20}
\rowcolor[RGB]{237,244,245}
SteerVTE (Ours)
& \textbf{0.9785}
& \textbf{0.9769}
& \textbf{0.9776}
& \textbf{0.9579}
& \textbf{0.7700}
& \textbf{0.9648}
& \textbf{35.55}
& \textbf{0.7200} \\
\bottomrule
\end{tabular}
}
% \vspace{-4mm}
\label{tab:realbench}
\end{table*}

\vspace{-5mm}

\begin{table*}[h]
\centering
\scriptsize
\setlength{\tabcolsep}{4.2pt}
\renewcommand{\arraystretch}{1.10}
\caption{\textbf{Quantitative comparison on VTE-Bench-SceneText.} Our method generalizes to unseen scene text scenarios. \textbf{Bold} indicates the best result. STRIVE* denotes our reproduced implementation.}
\vspace{2mm}
\resizebox{0.92\textwidth}{!}{
\begin{tabular}{l|ccccc|ccc}
\toprule
\textbf{Method}
& \multicolumn{5}{c|}{\textbf{Text Editing}}
& \multicolumn{3}{c}{\textbf{Video Quality}} \\
\cmidrule(lr){2-6} \cmidrule(lr){7-9}
& Precision$\uparrow$
& Recall$\uparrow$
& F-Score$\uparrow$
& NED$\uparrow$
& Accuracy$\uparrow$
& SSIM$_{bg}\uparrow$
& PSNR$_{bg}\uparrow$
& Quality Rate$\uparrow$ \\
\midrule
STRIVE*~\cite{subramanian2021strive}
& 0.5014
& 0.4943
& 0.4889
& 0.3834
& 0.0700
& \textbf{0.9794}
& \textbf{38.98}
& 0.0100 \\
TextCtrl~\cite{zeng2024textctrl}
& 0.7069
& 0.6670
& 0.6746
& 0.5927
& 0.3300
& \textbf{0.9794}
& \textbf{38.98}
& 0.0600 \\
AnyText2~\cite{tuo2024anytext2}
& 0.5727
& 0.5850
& 0.5636
& 0.4337
& 0.1000
& 0.8291
& 23.23
& 0.0100 \\
FLUX2~\cite{labs2025flux}
& 0.5581
& 0.5277
& 0.5332
& 0.4413
& 0.2000
& 0.6677
& 18.17
& 0.0700 \\
\rowcolor[RGB]{237,244,245}
SteerVTE (Ours)
& \textbf{0.7631}
& \textbf{0.7494}
& \textbf{0.7475}
& \textbf{0.6908}
& \textbf{0.5100}
& 0.9682
& 36.21
& \textbf{0.4100} \\
\bottomrule
\end{tabular}
}
\label{tab:scenetextbench}
\end{table*}

\subsection{Qualitative Evaluation}
\begin{figure}[h] % h表示here，即尽量放在当前位置
    \centering
    \includegraphics[width=1.0\textwidth]{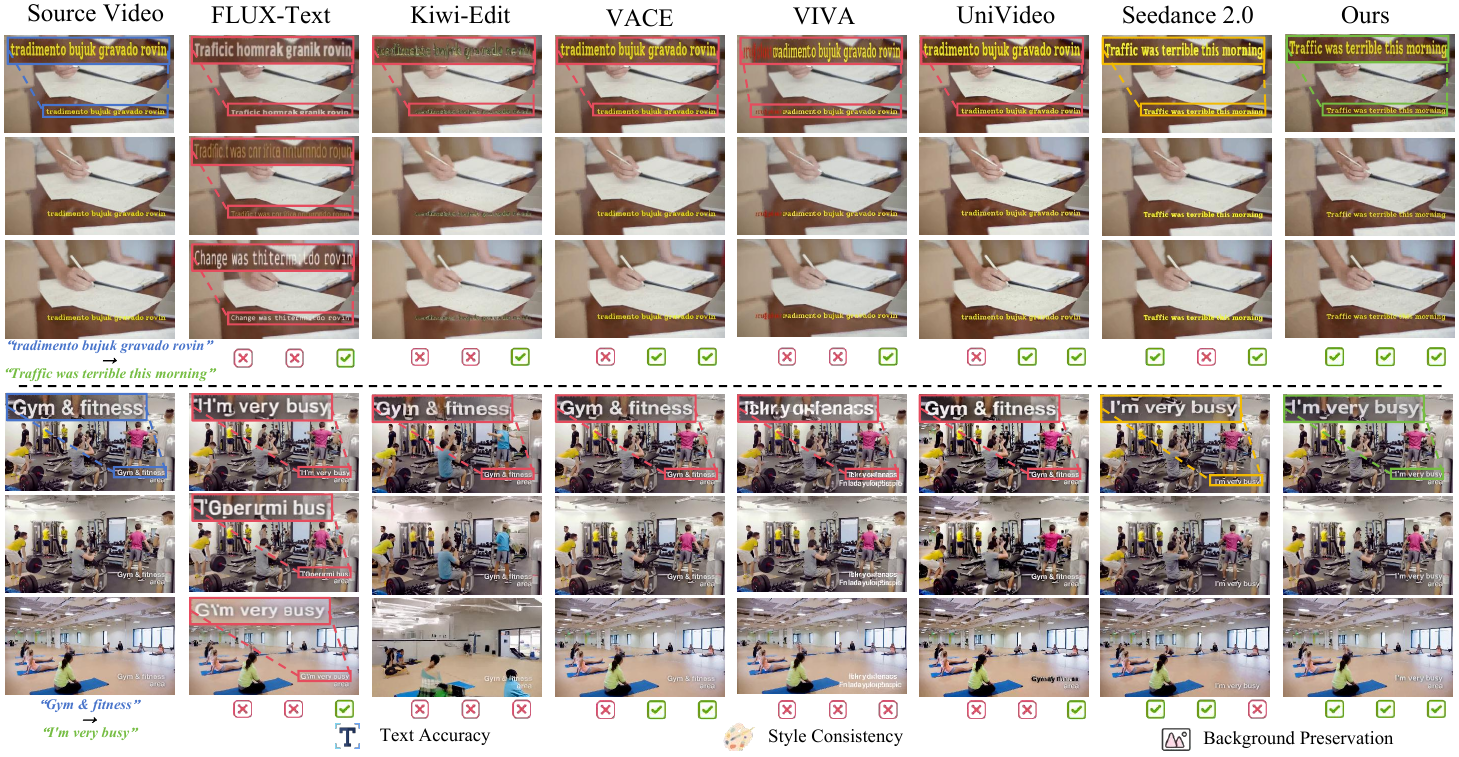} % 图片路径和大小
    % \vspace{-3mm}
    \caption{\textbf{Qualitative comparison.} Prior baselines fail in distinct ways: inaccurate or flickering edits (FLUX-Text), no edit at all (VACE), garbled or style-inconsistent glyphs (Kiwi-Edit, VIVA, UniVideo), or unintended modifications to background text (Seedance 2.0). Our method achieves seamless edits that preserve text accuracy, style consistency, and background integrity.}
    \vspace{-2mm}
    \label{fig:comparison}
\end{figure}

We conduct qualitative comparisons with state-of-the-art methods, including FLUX-Text, VACE, VIVA, UniVideo, Kiwi-Edit, and Seedance 2.0 on real-world scenarios (Fig.~\ref{fig:comparison}). Our method achieves precise text editing, consistent text style, and consistent background. Seedance 2.0 produces reasonable results but lacks perfect accuracy and makes erroneous background modifications. Other video editing models generally lack text editing capabilities: they either leave text unmodified or exhibit deficiencies in accuracy, style, and background consistency, sometimes even drastically altering the entire video. More examples of scene text editing results can be found in Appx.~\ref{appx:comparison}.

\subsection{Ablation Studies}
\subsubsection{Model Architecture}
We train each variant on the SynthTE-Hard video dataset and evaluate on VTE-Bench-Synth. Starting from VACE as the baseline, we gradually add components until reaching the full SteerVTE.

\begin{table*}[h]
\centering
\scriptsize
\setlength{\tabcolsep}{3.0pt}
\renewcommand{\arraystretch}{1.08}
\caption{\textbf{Ablation of model architecture.} We ablate input, glyph condition, and style encoder, selecting MV2V, dual-level glyph, and VLM for SteerVTE. Best results per sub-group are \textbf{bold}.}
\vspace{2mm}
\resizebox{\textwidth}{!}{
\begin{tabular}{cl|ccccc|ccccccc}
\toprule
\multicolumn{2}{c|}{\textbf{Architecture}}
& \multicolumn{5}{c|}{\textbf{Text Editing}}
& \multicolumn{7}{c}{\textbf{Video Quality}} \\
\cmidrule(lr){3-7} \cmidrule(lr){8-14}
&
& Precision$\uparrow$
& Recall$\uparrow$
& F-Score$\uparrow$
& NED$\uparrow$
& Accuracy$\uparrow$
& SSIM$\uparrow$
& PSNR$\uparrow$
& FID$\downarrow$
& FVD$\downarrow$
& Style-Sim$\uparrow$
& SSIM$_{bg}$$\uparrow$
& PSNR$_{bg}$$\uparrow$ \\
\midrule

\multirow{2}{*}{\rotatebox[origin=c]{90}{\textbf{Input}}}
& V2V
& 0.9340 & 0.8917 & 0.9123 & 0.7541 & 0.1300
& 0.2901 & \textbf{10.86} & 89.54 & 20.56 & 0.4847 & 0.9537 & 32.02 \\
& \cellcolor[RGB]{237,244,245}MV2V
& \cellcolor[RGB]{237,244,245}\textbf{0.9669}
& \cellcolor[RGB]{237,244,245}\textbf{0.9528}
& \cellcolor[RGB]{237,244,245}\textbf{0.9597}
& \cellcolor[RGB]{237,244,245}\textbf{0.8557}
& \cellcolor[RGB]{237,244,245}\textbf{0.2400}
& \cellcolor[RGB]{237,244,245}\textbf{0.3063}
& \cellcolor[RGB]{237,244,245}10.59
& \cellcolor[RGB]{237,244,245}\textbf{88.92}
& \cellcolor[RGB]{237,244,245}\textbf{18.98}
& \cellcolor[RGB]{237,244,245}\textbf{0.5029}
& \cellcolor[RGB]{237,244,245}\textbf{0.9617}
& \cellcolor[RGB]{237,244,245}\textbf{36.04} \\
\midrule

\multirow{3}{*}{\rotatebox[origin=c]{90}{\textbf{+Glyph}}}
& Line-Level
& 0.9788 & 0.9714 & 0.9750 & 0.9380 & 0.3900
& 0.3276 & 11.14 & 81.54 & \textbf{17.28} & 0.5041 & \textbf{0.9620} & 36.63 \\
& Character-Level
& 0.9659 & 0.9553 & 0.9605 & 0.8704 & 0.2700
& 0.3085 & 10.89 & 85.25 & 18.62 & 0.5050 & 0.9582 & 36.42 \\
& \cellcolor[RGB]{237,244,245}Dual-Level
& \cellcolor[RGB]{237,244,245}\textbf{0.9891}
& \cellcolor[RGB]{237,244,245}\textbf{0.9798}
& \cellcolor[RGB]{237,244,245}\textbf{0.9844}
& \cellcolor[RGB]{237,244,245}\textbf{0.9561}
& \cellcolor[RGB]{237,244,245}\textbf{0.5000}
& \cellcolor[RGB]{237,244,245}\textbf{0.3467}
& \cellcolor[RGB]{237,244,245}\textbf{11.74}
& \cellcolor[RGB]{237,244,245}\textbf{80.73}
& \cellcolor[RGB]{237,244,245}17.64
& \cellcolor[RGB]{237,244,245}\textbf{0.5163}
& \cellcolor[RGB]{237,244,245}0.9586
& \cellcolor[RGB]{237,244,245}\textbf{36.77} \\
\midrule

\multirow{3}{*}{\rotatebox[origin=c]{90}{\textbf{+Style}}}
& ViT
& 0.9665 & 0.9546 & 0.9605 & 0.9111 & 0.4000
& 0.2572 & 9.374 & 112.5 & 20.85 & 0.4387 & 0.9618 & 36.42 \\
& VLEmbedding
& \textbf{0.9889} & \textbf{0.9829} & \textbf{0.9858} & \textbf{0.9523} & 0.5100
& 0.3280 & 11.14 & 80.45 & \textbf{16.31} & 0.5256 & 0.9622 & 36.84 \\
& \cellcolor[RGB]{237,244,245}VLM
& \cellcolor[RGB]{237,244,245}0.9836
& \cellcolor[RGB]{237,244,245}0.9770
& \cellcolor[RGB]{237,244,245}0.9802
& \cellcolor[RGB]{237,244,245}0.9466
& \cellcolor[RGB]{237,244,245}\textbf{0.5300}
& \cellcolor[RGB]{237,244,245}\textbf{0.3562}
& \cellcolor[RGB]{237,244,245}\textbf{11.71}
& \cellcolor[RGB]{237,244,245}\textbf{79.59}
& \cellcolor[RGB]{237,244,245}16.90
& \cellcolor[RGB]{237,244,245}\textbf{0.5417}
& \cellcolor[RGB]{237,244,245}\textbf{0.9625}
& \cellcolor[RGB]{237,244,245}\textbf{37.06} \\
\bottomrule
\end{tabular}
}
\label{tab:ablation_architecture}
\end{table*}
\vspace{-2mm}

As shown in Tab.~\ref{tab:ablation_architecture}, the original V2V paradigm achieves only 13\% accuracy. Adding a mask improves accuracy by 11\% and enhances background preservation. Building on this mask-based V2V paradigm, we incorporate glyph prior information. Character-level glyph information raises accuracy to 27\%, line-level information raises it to 39\%, and using both granularities further boosts accuracy to 50\%. For the style encoder, we compare three variants: ViT, VLEmbedding, and VLM. ViT and VLM are derived from Qwen2.5-VL-3B~\cite{bai2025qwen3}, while VLEmbedding comes from Qwen3-VL-Embedding-2B~\cite{li2026qwen3}. ViT yields poor results, possibly due to a representation mismatch with DiT, whereas both VLEmbedding and VLM achieve strong improvements in accuracy and style consistency. We select VLM as the final style encoder due to its higher accuracy.

\subsubsection{Training Strategy}
To validate our three‑stage curriculum learning strategy, we conducted ablation studies on training data and losses. All experimental configurations were kept consistent with the main experiment. 
% \vspace{-8mm}
\vspace{-2mm}
\begin{table*}[h]
\centering
\scriptsize
\setlength{\tabcolsep}{3.1pt}
\renewcommand{\arraystretch}{1.08}
% \caption{\textbf{Ablation study on training strategy.} Top: ablation on training data, where Video-Only uses only video data, Hybrid Data mixes all training data in a single run, and Curriculum Learning adopts staged progression as described in Sec.~\ref{sec:training}. Bottom: ablation on training loss. Combining OCR loss and mask reweight loss achieves the best balance between text editing accuracy and video quality.}
\caption{\textbf{Ablation study on training data and loss.} Best results per setting are \textbf{bold}.}
\vspace{1mm}

\resizebox{\textwidth}{!}{
\begin{tabular}{l|ccccc|ccccccc}
\toprule
\multirow{2}{*}{\textbf{Training Data}}
% \textbf{Training Data}
& \multicolumn{5}{c|}{\textbf{Text Editing}}
& \multicolumn{7}{c}{\textbf{Video Quality}} \\
\cmidrule(lr){2-6} \cmidrule(lr){7-13}
& Precision$\uparrow$
& Recall$\uparrow$
& F-Score$\uparrow$
& NED$\uparrow$
& Accuracy$\uparrow$
& SSIM$\uparrow$
& PSNR$\uparrow$
& FID$\downarrow$
& FVD$\downarrow$
& Style-Sim$\uparrow$
& SSIM$_{bg}$$\uparrow$
& PSNR$_{bg}$$\uparrow$ \\
\midrule
Video-Only
& 0.8922
& 0.8925
& 0.8923
& 0.7563
& 0.3700
& 0.3371
& 11.73
& 84.67
& 17.88
& 0.5363
& 0.9589
& 36.58 \\
Hybrid Data
& \textbf{0.9916}
& 0.9783
& 0.9849
& 0.9530
& 0.5300
& 0.3467
& 10.69
& 83.67
& 15.49
& 0.5324
& 0.9523
& 35.92 \\
\rowcolor[RGB]{237,244,245}
Curriculum Learning
& 0.9908
& \textbf{0.9825}
& \textbf{0.9866}
& \textbf{0.9624}
& \textbf{0.6100}
& \textbf{0.4034}
& \textbf{12.30}
& \textbf{65.46}
& \textbf{14.55}
& \textbf{0.5936}
& \textbf{0.9606}
& \textbf{37.09} \\
\bottomrule
\end{tabular}
}

\vspace{2mm}

\resizebox{\textwidth}{!}{
\begin{tabular}{cc|ccccc|ccccccc}
\toprule
\multicolumn{2}{c|}{\textbf{Training Loss}}
% \textbf{Training Loss}
& \multicolumn{5}{c|}{\textbf{Text Editing}}
& \multicolumn{7}{c}{\textbf{Video Quality}} \\
\cmidrule(lr){1-2} \cmidrule(lr){3-7} \cmidrule(lr){8-14}
OCR & SF
& Precision$\uparrow$
& Recall$\uparrow$
& F-Score$\uparrow$
& NED$\uparrow$
& Accuracy$\uparrow$
& SSIM$\uparrow$
& PSNR$\uparrow$
& FID$\downarrow$
& FVD$\downarrow$
& Style-Sim$\uparrow$
& SSIM$_{bg}$$\uparrow$
& PSNR$_{bg}$$\uparrow$ \\
\midrule
$\times$ & $\times$
& 0.9848
& 0.9774
& 0.9810
& 0.9474
& 0.5100
& 0.3481
& 11.24
& 79.26
& 16.44
& 0.5268
& 0.9604
& 35.51 \\
$\checkmark$ & $\times$
& 0.9928
& \textbf{0.9902}
& \textbf{0.9914}
& \textbf{0.9751}
& \textbf{0.6500}
& 0.2881
& 10.04
& 108.1
& 18.03
& 0.4794
& 0.9605
& 36.09 \\
$\times$ & $\checkmark$
& \textbf{0.9929}
& 0.9817
& 0.9872
& 0.9613
& 0.5600
& 0.3756
& 11.33
& 69.84
& \textbf{13.99}
& 0.5691
& \textbf{0.9608}
& 35.71 \\
\rowcolor[RGB]{237,244,245}
$\checkmark$ & $\checkmark$
& 0.9908
& 0.9825
& 0.9866
& 0.9624
& 0.6100
& \textbf{0.4034}
& \textbf{12.30}
& \textbf{65.46}
& 14.55
& \textbf{0.5936}
& 0.9606
& \textbf{37.09} \\
\bottomrule
\end{tabular}
}
\label{tab:ablation_training}
\end{table*}

\textbf{Training Data.}
As shown at the top of Tab.~\ref{tab:ablation_training}, the \emph{Video-Only} setup, which uses only video data without pretraining on image data, leads to degraded performance across all metrics, confirming that video foundation models have weak text priors and benefit from training data organized by increasing difficulty. \emph{Hybrid Data}, which mixes all data in a single training stage, improves text accuracy but yields suboptimal results due to conflicts between image and video characteristics. Our three-stage \emph{Curriculum Learning} first uses image data for basic text editing, then video data for temporal consistency, achieving optimal results on nearly all metrics.

\textbf{Training Loss.}
We conduct this experiment during the third training stage using flow matching loss as the baseline. As shown at the bottom of Tab.~\ref{tab:ablation_training}, the baseline struggles with both text accuracy and style consistency. Adding a Glyph-Aware OCR loss raises text accuracy by 14 points but degrades style consistency by biasing generation toward simple fonts and colors. Adding a Spatial-Focal Loss instead concentrates supervision on the editing region, improving style consistency with little gain in text accuracy.
To combine the best of both worlds, GLAS Loss amalgamates the two losses, striking an optimal balance between text accuracy and style fidelity.
% Owing to their complementary nature, GLAS Loss amalgamates the two losses, striking an optimal balance between text accuracy and style fidelity.
\vspace{-2mm}
\section{Conclusion}
\vspace{-2mm}

This work introduces SteerVTE, the first end-to-end architecture for seamless video text editing. At its core, a frozen DiT backbone is augmented with three precise editing signals: a spatiotemporal mask that localizes the target region, a VLM-based style encoder that captures fine-grained visual attributes, and a dual-granularity glyph module that fuses character- and sentence-level representations. To overcome the inherently weak glyph priors of video foundation models, we design a three-stage curriculum that progresses from elementary glyph learning, through complex real-world text, and finally to full temporal-coherence training. We further propose a GLAS loss that concentrates supervision on text regions, boosting sentence-level fidelity and style consistency. To support training and evaluation, we construct SteerVTE-1M, a million-triplet dataset spanning diverse scenes and text, and VTE-Bench, the first dedicated benchmark for video text editing. 

% \textbf{Limitations and Future Work.} SteerVTE currently supports only English and is designed for single-line editing. To overcome these limitations, future work will extend it to other languages using larger multilingual datasets and enable multi-line editing through improved foundation models. Beyond font diversity, dynamic subtitles with richer text effects are also promising. While our model achieves over 95\% normalized edit distance, accuracy remains paramount since even single-character errors are often unacceptable. Reinforcement learning, such as GRPO, could help close the remaining gap. Model distillation further offers a path toward lightweight real-time subtitle editing.

\textbf{Limitations and Future Work.} SteerVTE currently supports only English and is designed for single-line editing. To overcome these limitations, future work will extend it to other languages using larger multilingual datasets and enable multi-line editing through improved foundation models. Beyond font diversity, dynamic subtitles with richer text effects are also promising. While our model achieves over 95\% normalized edit distance, accuracy remains paramount since even single-character errors are often unacceptable. Reinforcement learning, such as GRPO, could help close the remaining gap. Model distillation further offers a path toward lightweight real-time subtitle editing.

Limited by the capabilities of video foundation models, long-video text editing remains a limitation of our approach. Future work could address this through more efficient model architectures or algorithms that improve long-context modeling. More importantly, real-world video text editing data remain scarce, especially paired scene text editing videos that cover cases where the text moves with the scene, or where the text is deformed, zoomed, and otherwise transformed. Training on such data is expected to yield better results than the hybrid image-video training data we currently use.
\section{Acknowledgments}
We would like to thank Yiming Zhao and Siyu Jiao for many insightful discussions.
This work is supported by Beijing Municipal Natural Science Foundation (Grant No. L2603008), Fundamental and Interdisciplinary Disciplines Breakthrough Plan of the Ministry of Education of China (JYB2025XDXM113), National Natural Science Foundation of China (92470121, 62402016), National Key R\&D Program of China (2024YFA1014003), Zhongguancun Academy (C20250204, C20250602),  Beijing Major Science and Technology Project (Z251100008125043, Z251100008425023), and High-performance Computing Platform of Peking University.

Competing Interests. The authors declare no competing interests beyond the affiliations and funding sources disclosed in this paper.
% \inputsection{secs1_abstract}
% \inputsection{2_introduction}
% \inputsection{3_related_work}
% \inputsection{4_method}
% \inputsection{5_results}
% \inputsection{6_conclusions}
% \inputsection{8_acknowledgement}

\newpage
\bibliographystyle{plain}
\bibliography{papers}

%%%%%%%%%%%%%%%%%%%%%%%%%%%%%%%%%%%%%%%%%%%%%%%%%%%%%%%%%%%%

\newpage
\appendix
\section{Outlines}
The supplementary material presents the following sections to strengthen the main manuscript:
\begin{itemize}[leftmargin=1em]
\item Section~\ref{appx:dataset} presents the complete construction pipeline of our SteerVTE-1M dataset and Section~\ref{appx:loss_details} includes a detailed derivation of the GLAS Loss.
\item Section~\ref{appx:benchmark} introduces the composition of our VTE-Bench evaluation suite and the specifics of each evaluation metric, while Section~\ref{appx:implementation} elaborates on training and inference details.
\item Section~\ref{appx:quantitative} presents extra quantitative comparisons with existing image editing methods.
\item Section~\ref{appx:ablation} provides additional visualizations of our ablation studies.
\item Section~\ref{appx:comparison} presents extra qualitative comparisons with existing video editing methods.
\end{itemize}

\section{Additional Method Details}
\label{appx:method}
\subsection{Dataset Construction}
\label{appx:dataset}
\begin{figure}[h]
    \centering
    \includegraphics[width=1.0\textwidth]{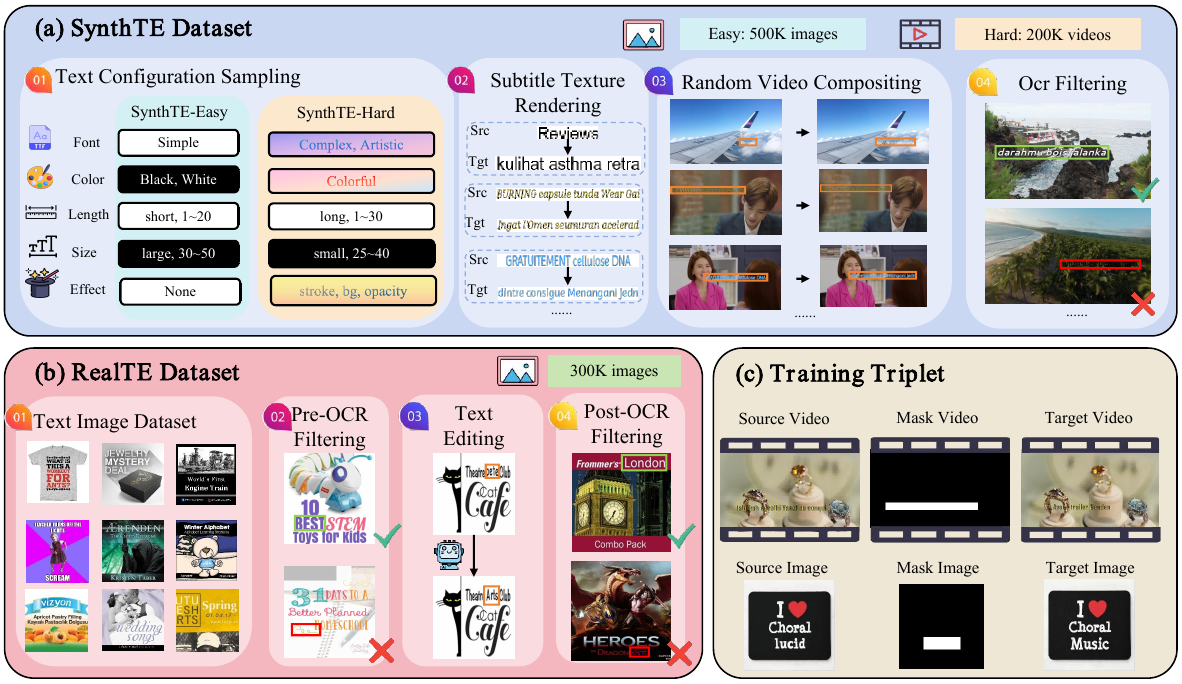}
    \caption{\textbf{Construction process of the visual text editing training dataset.} (a) Using the same procedure with two different text configurations, we synthesize the SynthTE-Easy and SynthTE-Hard. (b) We construct the image text editing RealTE dataset from real-world OCR data. (c) Example samples from the video and image datasets.}
    \label{fig:dataset_app}
\end{figure}

This appendix provides a comprehensive description of the SteerVTE-1M dataset construction, complementing the summary in Sec.~\ref{sec:dataset}. The dataset comprises two complementary components: SynthTE, a stylized synthetic dataset with diverse scenes, fonts, and effects, and RealTE, a real-world image-based text editing dataset. Both come with fine-grained annotations, including text strings, bounding boxes, and editing region masks. Fig.~\ref{fig:dataset_app} illustrates the construction pipeline.

For SynthTE, we collect 1,933 TrueType fonts from Google Fonts~\cite{google_fonts}, an English text corpus~\cite{google_10000_english} of 65 letters and symbols, and background videos from opensora-plan~\cite{lin2024open}. Given the weak text rendering of video foundation models, we split SynthTE into easy and hard versions to support different training stages. The easy version uses 894 simple black-and-white fonts, restricts text length to under 20 characters, and sets the font size from 30 to 50. The hard version uses all available fonts, including artistic and textured styles, randomly samples colors, and adds transparency, background boxes, and strokes, with text length up to 30 and font size from 25 to 40.

As illustrated in Fig.~\ref{fig:dataset_app}(a), the pipeline randomly samples text attributes and paired text to render subtitle images, selects background videos, and generates bounding boxes covering 8\% to 15\% of the frame for pasting. It inserts one to four subtitle pairs, selects one as the training region, and retains only samples that achieve 100\% OCR recognition accuracy. This process produces 500,000 images for SynthTE-Easy and 200,000 videos for SynthTE-Hard.

For RealTE, we note that SynthTE alone cannot address real-world scenarios, yet real video text editing data is scarce. Since no existing model preserves style and temporal consistency when editing videos, paired data is unavailable. We therefore curated a real image-based text editing dataset from design images (Fig.~\ref{fig:dataset_app}(b)), leveraging OCR-annotated images and image text editors. Our experiments show that editing capabilities learned from images generalize to videos. From AnyWord-3M~\cite{tuo2023anytext}, we used an OCR model~\cite{cui2025paddleocr} to retain only fully recognizable editing regions, yielding 500k samples. We then applied a text editing model~\cite{zeng2024textctrl} and OCR filtering to ensure clear text before and after editing. To preserve stylistic consistency, we also employ a Text Style Similarity metric~\cite{seo2025stellar} and discard samples with scores below 0.4. 
This pipeline produces the RealTE Dataset with 300k image samples.
% producing the RealTE Dataset with 300k image samples.

\begin{figure}[h] % h表示here，即尽量放在当前位置
    \centering
    \includegraphics[width=0.84\textwidth]{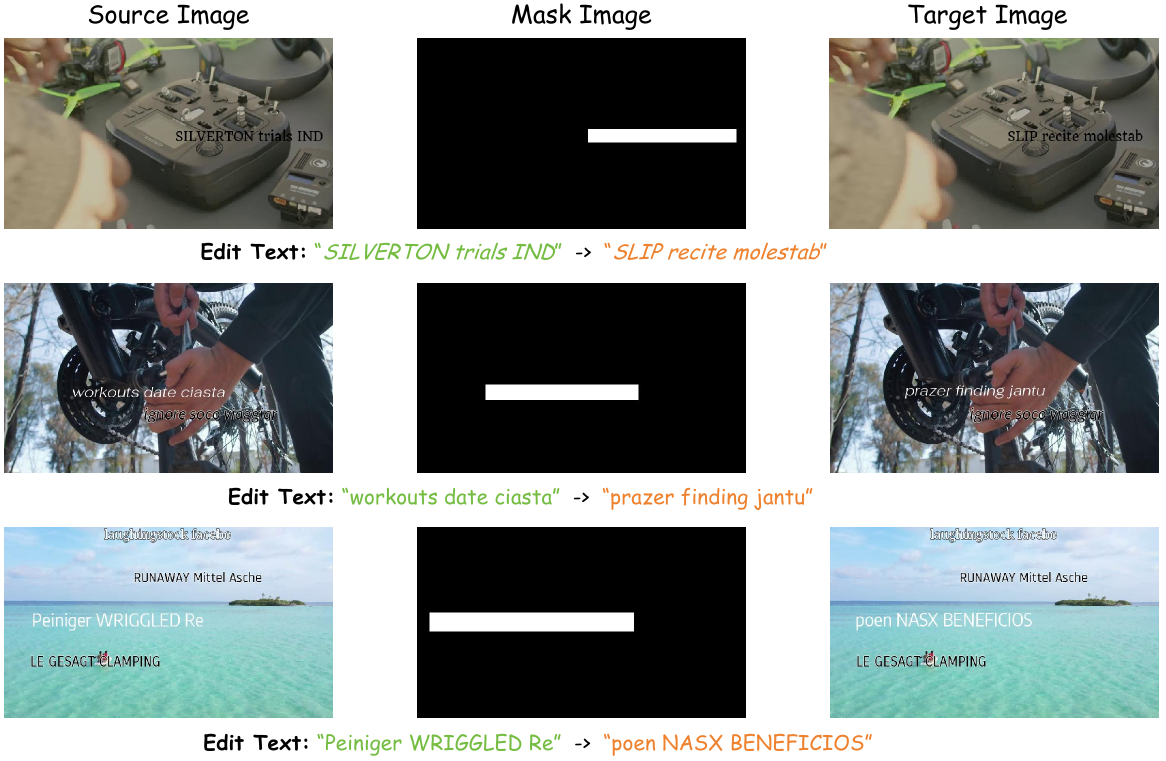} % 图片路径和大小
    % \vspace{-3mm}
    \caption{Examples from our stage-1 image training dataset: SynthTE-Easy.}
    \vspace{-2mm}
    \label{fig:appendix_dataset_1}
\end{figure}

\begin{figure}[h] % h表示here，即尽量放在当前位置
    \centering
    \includegraphics[width=0.7\textwidth]{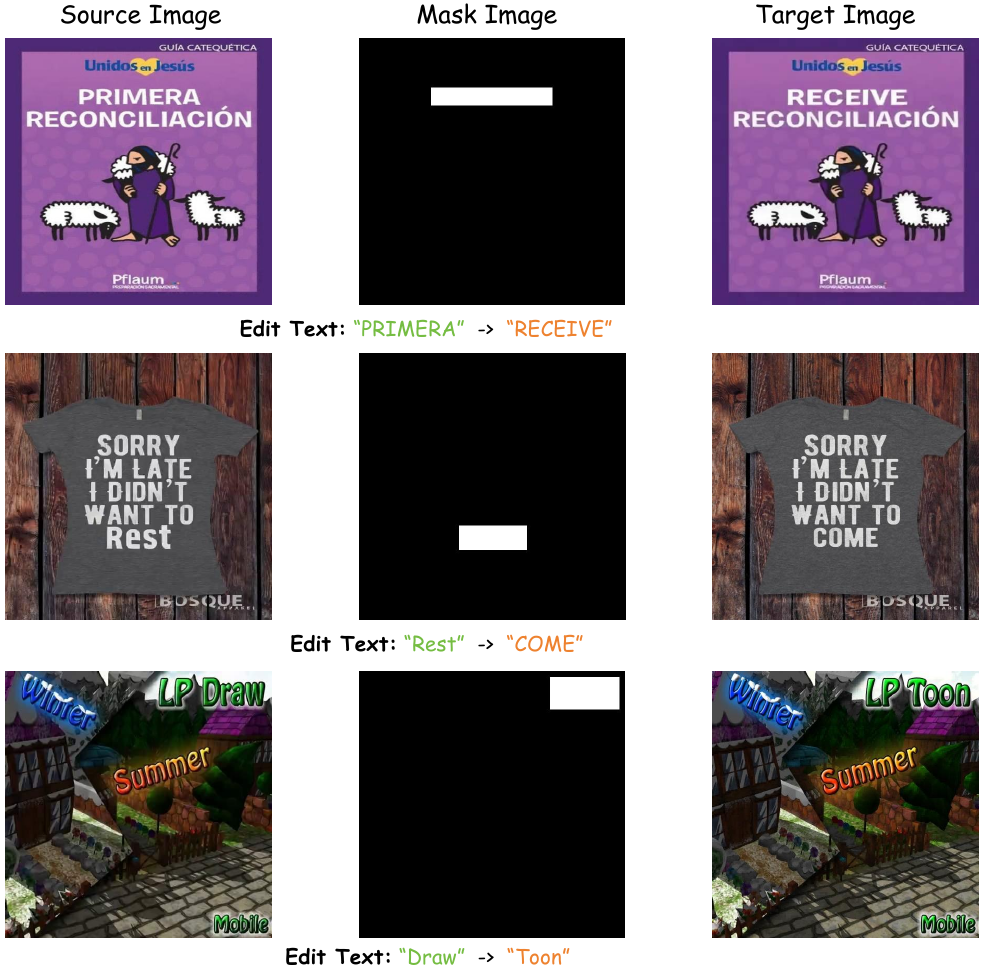} % 图片路径和大小
    % \vspace{-3mm}
    \caption{Examples from our stage-2 image training dataset: RealTE.}
    \vspace{-2mm}
    \label{fig:appendix_dataset_3}
\end{figure}

\begin{figure}[h] % h表示here，即尽量放在当前位置
    \centering
    \includegraphics[width=0.84\textwidth]{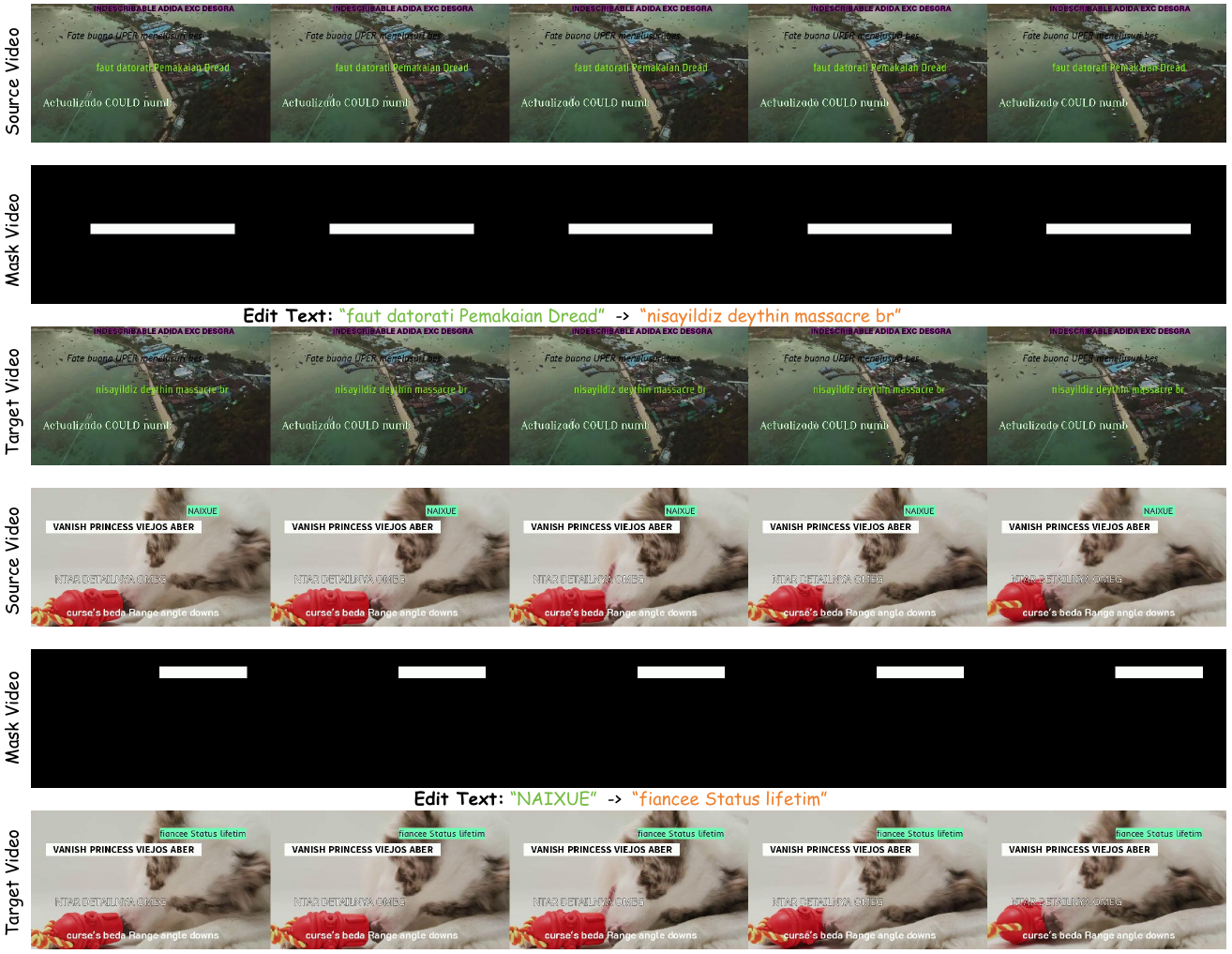} % 图片路径和大小
    % \vspace{-3mm}
    \caption{Examples from our stage-3 video training dataset: SynthTE-Hard.}
    \vspace{-2mm}
    \label{fig:appendix_dataset_6}
\end{figure}

We present additional examples from our diverse SteerVTE-1M dataset, including samples from all three training stages (SynthTE-Easy, RealTE, and SynthTE-Hard), as illustrated in Figures~\ref{fig:appendix_dataset_1}--\ref{fig:appendix_dataset_6}. For a better visual experience, please visit our anonymous website \url{https://steervte-paper.github.io/}, which includes video demonstrations of qualitative comparisons and dataset samples.

\subsection{GLAS Loss Details}
\label{appx:loss_details}

We provide the detailed formulation of the OCR supervision term in GLAS. Given the model prediction at timestep $t$, we first recover the clean latent estimate $\hat{\mathbf{z}}_0$ and decode it with the VAE decoder $\mathcal{D}$ to obtain the reconstructed video
\begin{equation}
\hat{\mathbf{V}} = \mathcal{D}(\hat{\mathbf{z}}_0).
\end{equation}
We then uniformly sample aligned frame triplets from the ground-truth video, reconstructed video, and mask video:
\begin{equation}
(\mathbf{I}_{\mathrm{gt}}, \mathbf{I}_{\mathrm{pred}}, \mathbf{I}_{\mathrm{m}})
=
\mathcal{S}\!\left(\mathbf{V}_{\mathrm{gt}},\, \hat{\mathbf{V}},\, \mathbf{M}\right).
\end{equation}

We use a pre-trained OCR recognizer $\mathcal{O}$~\cite{cui2025paddleocr} and distinguish its outputs for different purposes. For an input text patch $\mathbf{x}$, the recognizer produces
\begin{equation}
\bigl\{
\mathcal{O}_{\mathrm{neck}}(\mathbf{x}),\;
\mathcal{O}_{\mathrm{feat}}(\mathbf{x}),\;
\mathcal{O}_{\mathrm{logits}}(\mathbf{x})
\bigr\},
\end{equation}
where $\mathcal{O}_{\mathrm{neck}}$ denotes the neck feature used in the main text for cross-attention injection, $\mathcal{O}_{\mathrm{feat}}$ denotes the multi-scale feature maps used in $\mathcal{L}_{\text{text}}$, and $\mathcal{O}_{\mathrm{logits}}$ denotes the character logits used in $\mathcal{L}_{\text{CTC}}$.

Given the sampled mask $\mathbf{I}_{\mathrm{m}}$, we crop the text-bearing regions from $\mathbf{I}_{\mathrm{gt}}$ and $\mathbf{I}_{\mathrm{pred}}$, and normalize them with the OCR preprocessing pipeline:
\begin{equation}
\tilde{\mathbf{P}}_{\mathrm{gt}}=\mathcal{N}\!\left(\mathrm{Crop}(\mathbf{I}_{\mathrm{gt}}, \Omega(\mathbf{I}_{\mathrm{m}}))\right),\quad
\tilde{\mathbf{P}}_{\mathrm{pred}}=\mathcal{N}\!\left(\mathrm{Crop}(\mathbf{I}_{\mathrm{pred}}, \Omega(\mathbf{I}_{\mathrm{m}}))\right),
\end{equation}
where $\Omega(\mathbf{I}_{\mathrm{m}})$ denotes the text region indicated by the mask.

\paragraph{Text perceptual loss.}
To explicitly constrain glyph topology, we match OCR feature maps between the predicted and target text patches. Let
\begin{equation}
\mathcal{O}_{\mathrm{feat}}(\tilde{\mathbf{P}}_{\mathrm{gt}})
=
\{\mathbf{F}^{(l)}_{\mathrm{gt}}\}_{l=1}^{L},\qquad
\mathcal{O}_{\mathrm{feat}}(\tilde{\mathbf{P}}_{\mathrm{pred}})
=
\{\mathbf{F}^{(l)}_{\mathrm{pred}}\}_{l=1}^{L},
\end{equation}
where $\mathbf{F}^{(l)}_{\mathrm{gt}}, \mathbf{F}^{(l)}_{\mathrm{pred}} \in \mathbb{R}^{H_l \times W_l \times C_l}$. We define
\begin{equation}
\mathcal{L}_{\text{text}}
=
\sum_{l=1}^{L}
\frac{1}{H_l W_l C_l}
\left\|
\mathbf{F}^{(l)}_{\mathrm{pred}}-\mathbf{F}^{(l)}_{\mathrm{gt}}
\right\|_2^2.
\end{equation}

\paragraph{CTC recognition loss.}
To further ensure semantic correctness, we apply a CTC loss on the OCR character logits:
\begin{equation}
\mathbf{Y}_{\mathrm{pred}}
=
\mathcal{O}_{\mathrm{logits}}(\tilde{\mathbf{P}}_{\mathrm{pred}})
\in \mathbb{R}^{T \times |\mathcal{V}|},
\end{equation}
\begin{equation}
\mathcal{L}_{\text{CTC}}
=
\mathrm{CTC}(\mathbf{Y}_{\mathrm{pred}}, y)
=
-\log p(y \mid \mathbf{Y}_{\mathrm{pred}}),
\end{equation}
where $y$ is the target transcription.

The OCR supervision term is thus
\begin{equation}
\mathcal{L}_{\text{OCR}}=\mathcal{L}_{\text{text}}+\mathcal{L}_{\text{CTC}}.
\end{equation}

It is worth noting that, to ensure the effectiveness of the OCR loss computation and training stability, we only apply the OCR loss at low-noise timesteps (\(t < 0.3\)). At these timesteps, the predicted latent variable \(\hat{\mathbf{z}}_0\) is more likely to decode into a clear image, preventing blurry and artifact-ridden frames from causing the OCR model to produce random and meaningless gradients.

For completeness, the latent-space spatial-focal loss in the main text is
\begin{equation}
\mathcal{L}_{\text{SF}}
=
\left\|
\mathbf{W}\odot
\left(
v_{\theta}(\cdot)-(\mathbf{z}_1-\mathbf{z}_0)
\right)
\right\|_2^2,
\end{equation}
where $\mathbf{W}$ is the normalized focal map derived from the text mask $\mathbf{M}$.

Finally, the overall GLAS objective is
\begin{equation}
\mathcal{L}_{\text{GLAS}}
=
\gamma \mathcal{L}_{\text{SF}}+\lambda \mathcal{L}_{\text{OCR}},
\end{equation}
where we use $\gamma=3.0$ and $\lambda=0.1$ in all experiments.

\section{Additional Experiment Details}
\label{appx:experiment}
% In this section, we describe the input configurations and hyperparameters used for all methods.

\subsection{Benchmark and Metrics}
\label{appx:benchmark}
\textbf{Construction details.}  
For VTE-Bench-Synth, we synthesize 50 test samples from SynthTE-Easy and 50 from SynthTE-Hard following the procedure described in Sec.~\ref{sec:dataset}. Each sample contains a source video, a random editing mask, an editing prompt, and the ground-truth edited video.  
For VTE-Bench-Real, we build 100 test samples from real-world video text detection datasets~\cite{wu2021bilingual} without ground-truth edited videos.
For VTE-Bench-SceneText, we build 200 test samples from STRIVE~\cite{subramanian2021strive} and the web, covering static, dynamic, and deformable scene text. This subset represents the most challenging setting for video text editing.

% The training videos used in this paper are sourced from ~ \url{https://github.com/PKU-YuanGroup/Open-Sora-Plan}. The original authors permit academic use, and the selected videos are released under the MIT License. The test videos are sourced from \url{https://github.com/weijiawu/BOVText-Benchmark}. The original authors permit academic use, and the selected videos are released under the Creative Commons Attribution 4.0 International License.
The training videos used in this paper are sourced from this~\cite{lin2024open}. The original authors permit academic use, and the selected videos are released under the MIT License. The test videos are sourced from this ~\cite{wu2021bilingual}. The original authors permit academic use, and the selected videos are released under the Creative Commons Attribution 4.0 International License.

\textbf{Metric computation details.}  
Except for FVD~\cite{unterthiner2019fvd}, all metrics are computed by uniformly sampling five frames from each video and averaging the results across frames.  
For text editing accuracy, we use PP-OCRv4~\cite{cui2025paddleocr} for text detection and recognition. Our SteerVTE model employs PP-OCRv3 as the glyph encoder and for OCR-loss computation due to its more accessible architecture, while the stronger PP-OCRv4 is adopted solely for evaluation to ensure fairness. Based on the OCR predictions and ground-truth texts, we report precision, recall, F-score, normalized edit distance (NED), and sentence accuracy.  
A higher NED indicates better performance:
\[
\mathrm{NED} = 1 - \frac{\mathrm{ED}(\mathrm{pred}, \mathrm{gt})}{\max(\mathrm{len}(\mathrm{pred}), \mathrm{len}(\mathrm{gt}))},
\]
where \(\mathrm{ED}\) is the Levenshtein edit distance between the OCR output \(\mathrm{pred}\) and the ground-truth text \(\mathrm{gt}\). Sentence accuracy is defined as 1 if the full OCR string exactly matches the target, and 0 otherwise. Sentence accuracy is stricter than NED, since any single character error causes the sentence to fail.

For video quality assessment, we use SSIM~\cite{wang2004image}, PSNR~\cite{PSNR_Wikipedia_1210897995}, and FID~\cite{heusel2017gans} to measure visual similarity between the ground-truth and edited frames; Text Style Similarity~\cite{seo2025stellar} (denoted as Style-Sim) to evaluate style consistency; and FVD to measure temporal coherence. Since the edited regions occupy only 10\%--15\% of each video, these metrics are computed on the cropped edited regions. To further assess background preservation, we compute SSIM and PSNR on frames with the edited regions masked out, denoted as SSIM$_{bg}$ and PSNR$_{bg}$, respectively.  
As VTE-Bench-Real does not provide ground-truth edited videos, we additionally conduct a user study involving twenty-five participants from varied backgrounds (non-researchers, industry, and students). Each participant assigns a binary score to each result: 1 if the edit contains accurate text, preserves style consistency, and leaves the background unchanged, and 0 otherwise. The final Quality Rate is computed as the average of these scores. Moreover, since SSIM$_{bg}$ and PSNR$_{bg}$ do not require edited-video ground truth, we compute them directly between the edited videos and the original input videos.

\subsection{Implementation Details}
\label{appx:implementation}
\textbf{More implementation details.} Stage 1 uses SynthTE Easy with 500k image samples at a resolution of $832 \times 480$ for 40k optimization steps. Stage 2 uses RealTE with 300k image samples at a resolution of $512 \times 512$ for 20k steps. Stage 3 uses SynthTE Hard with 200k video samples at a resolution of $832 \times 480$ with 49 frames per video for 23k steps. 
% The OGR loss hyperparameters are set as $\alpha = 1$, $\beta = 1$, $\gamma = 3$, and $\lambda = 0.1$. 
All training experiments are conducted on 16 NVIDIA H100 GPUs with a learning rate of $1 \times 10^{-5}$ using the AdamW optimizer and a batch size of 1. The resolution of the line-level glyph image matches the original image resolution, while the resolution of the character-level glyph image is $80 \times 80$. 

\textbf{Model Input.} For our method, the model takes as input the source video, an editing mask video, a text prompt, and both line-level and character-level glyph images rendered from the target text. The style reference image is obtained by extracting the masked text region from the first frame of the source video. The prompt for the style encoder is: ``Describe in detail the typography, color, style, text material, and rendering effects of the text regions \{source text\} in this image.'' The prompt for the T5 encoder is: ``A clear and high-quality video with smooth motion, featuring the stylized text: \{target text\}.'' For other mask-based video editing baselines, the input consists of the source video, the editing mask video, a line-level glyph image, and a text prompt. To explicitly indicate the editable region, the prompt is set to: ``Change the text from \{source text\} to \{target text\}, while keeping the text style consistent.'' For instruction-guided models, reference-guided models, and Seedance 2.0, which do not support editing masks, the input includes the source video, a line-level glyph image, and the same text prompt used for mask-based video editing models.

\begin{wraptable}{r}{0.42\textwidth}
\vspace{-1.3em}
\centering
\caption{Average inference time on a 49-frame video with resolution $832 \times 480$.}
\label{tab:inference_speed}
\begin{tabular}{lc}
\toprule
\textbf{Method} & \textbf{Time (s)} \\
\midrule
VACE~\cite{jiang2025vace} (14B) & 103 \\
VideoPainter~\cite{bian2025videopainter} (12B + 5B) & 81 \\
VIVA~\cite{cong2025viva} (8B + 13B) & 161 \\
UniVideo~\cite{wei2025univideo} (7B + 13B) & 277 \\
Kiwi-Edit~\cite{lin2026kiwi} (3B + 5B) & 8 \\
Seedance 2.0~\cite{seedance2026seedance} & 126 \\
Ours (3B + 14B) & 108 \\
\bottomrule
\end{tabular}
\vspace{-1.2em}
\end{wraptable}

\textbf{Inference speed.} We report the average inference time of each model on a single 49-frame video at a resolution of $832 \times 480$. As shown in Table~\ref{tab:inference_speed}, Kiwi-Edit is the fastest method, while our method achieves competitive efficiency comparable to other 14B-class baselines.

\textbf{Note.} All open-source methods, including ours, use 20 inference steps, since we empirically observe negligible differences compared with 50 steps. For open-source baselines, we use their official implementations. For Seedance 2.0, we use the official 480p API.

\section{Additional Quantitative Evaluation}
\label{appx:quantitative}
We additionally compare with STRIVE~\cite{subramanian2021strive}, TextCtrl~\cite{zeng2024textctrl}, AnyText2~\cite{tuo2024anytext2}, and FLUX2~\cite{labs2025flux} on VTE-Bench-Synth and VTE-Bench-Real. As shown in Tab.~\ref{tab:synbench_additional} and Tab.~\ref{tab:realbench_additional}, image text editors struggle with sentence-level correctness and temporal stability in videos. In contrast, SteerVTE directly edits videos, achieving higher text accuracy, better style consistency, and more stable results. STRIVE and TextCtrl edit only foreground text regions and reuse the original backgrounds, which explains their higher background scores.

\begin{table*}[h]
\centering
\scriptsize
\setlength{\tabcolsep}{3.2pt}
\renewcommand{\arraystretch}{1.10}
\vspace{-4mm}
\caption{\textbf{Quantitative comparison on VTE-Bench-Synth.} Comparison with additional text editing methods. \textbf{Bold} is the best. STRIVE* is our reproduced implementation.}
\vspace{2mm}
\resizebox{\textwidth}{!}{
\begin{tabular}{l|ccccc|ccccccc}
\toprule
\textbf{Method}
& \multicolumn{5}{c|}{\textbf{Text Editing}}
& \multicolumn{7}{c}{\textbf{Video Quality}} \\
\cmidrule(lr){2-6} \cmidrule(lr){7-13}
& Precision$\uparrow$
& Recall$\uparrow$
& F-Score$\uparrow$
& NED$\uparrow$
& Accuracy$\uparrow$
& SSIM$\uparrow$
& PSNR$\uparrow$
& FID$\downarrow$
& FVD$\downarrow$
& Style-Sim$\uparrow$
& SSIM$_{bg}$$\uparrow$
& PSNR$_{bg}$$\uparrow$ \\
\midrule
STRIVE$^*$
& 0.8574
& 0.8135
& 0.8271
& 0.6913
& 0.0400
& 0.2895
& 12.47
& 107.1
& 16.49
& 0.5199
& \textbf{0.9773}
& \textbf{39.85} \\
TextCtrl
& 0.8176
& 0.5832
& 0.6559
& 0.3700
& 0.0100
& 0.3430
& \textbf{12.69}
& 98.10
& 127.8
& 0.5336
& \textbf{0.9773}
& \textbf{39.85} \\
AnyText2
& 0.7393
& 0.6516
& 0.6803
& 0.3999
& 0.0300
& 0.1832
& 10.97
& 163.4
& 884.2
& 0.4056
& 0.8197
& 23.42 \\
FLUX2
& 0.7707
& 0.6736
& 0.7088
& 0.5706
& 0.0800
& 0.2333
& 11.37
& 108.7
& 372.2
& 0.4538
& 0.7031
& 22.22 \\
\rowcolor[RGB]{237,244,245}
SteerVTE (Ours)
& \textbf{0.9908}
& \textbf{0.9825}
& \textbf{0.9866}
& \textbf{0.9624}
& \textbf{0.6100}
& \textbf{0.4034}
& 12.30
& \textbf{65.46}
& \textbf{14.55}
& \textbf{0.5936}
& 0.9606
& 37.09 \\
\bottomrule
\end{tabular}
}
\label{tab:synbench_additional}
\end{table*}

\begin{table*}[h]
\centering
\scriptsize
\setlength{\tabcolsep}{4.2pt}
\renewcommand{\arraystretch}{1.10}
\caption{\textbf{Quantitative comparison on VTE-Bench-Real.} Comparison with additional text editing methods. SteerVTE achieves the best text editing performance. \textbf{Bold} is the best.}
\vspace{2mm}
\resizebox{0.92\textwidth}{!}{
\begin{tabular}{l|ccccc|cc}
\toprule
\textbf{Method}
& \multicolumn{5}{c|}{\textbf{Text Editing}}
& \multicolumn{2}{c}{\textbf{Video Quality}} \\
\cmidrule(lr){2-6} \cmidrule(lr){7-8}
& Precision$\uparrow$
& Recall$\uparrow$
& F-Score$\uparrow$
& NED$\uparrow$
& Accuracy$\uparrow$
& SSIM$_{bg}$$\uparrow$
& PSNR$_{bg}$$\uparrow$ \\
\midrule
STRIVE$^*$
& 0.6900
& 0.6754
& 0.6752
& 0.6005
& 0.1300
& \textbf{0.9809}
& \textbf{40.35} \\
TextCtrl
& 0.8476
& 0.7855
& 0.8088
& 0.6702
& 0.2700
& \textbf{0.9809}
& \textbf{40.35} \\
AnyText2
& 0.7058
& 0.7061
& 0.6968
& 0.5300
& 0.0800
& 0.8473
& 22.11 \\
FLUX2
& 0.6538
& 0.5904
& 0.6082
& 0.4652
& 0.0900
& 0.6860
& 18.24 \\
\rowcolor[RGB]{237,244,245}
SteerVTE (Ours)
& \textbf{0.9785}
& \textbf{0.9769}
& \textbf{0.9776}
& \textbf{0.9579}
& \textbf{0.7700}
& 0.9648
& 35.55 \\
\bottomrule
\end{tabular}
}
\label{tab:realbench_additional}
\end{table*}

\section{Additional Ablation Study}
\label{appx:ablation}
\begin{figure}[h!]
    \centering
    \includegraphics[width=1.0\textwidth]{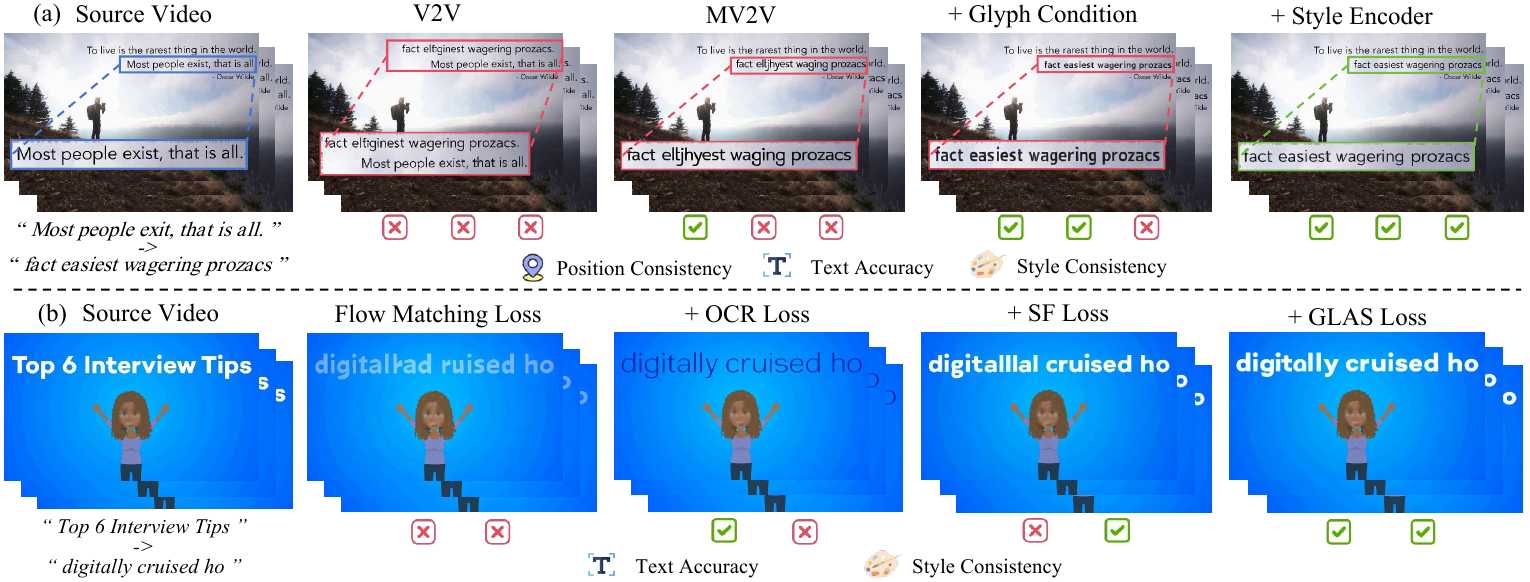} % 图片路径和大小
    \vspace{-3mm}
    \caption{\textbf{Visualization of ablation experiments.} (a) Model architecture: MV2V enables accurate region localization; glyph conditioning improves text accuracy; and the style encoder ensures style consistency. (b) Training loss: Using either the OCR loss or the SF loss alone yields suboptimal results. The GLAS loss achieves both text accuracy and style consistency.}
    \vspace{-5mm}
    \label{fig:abla}
\end{figure}

As shown in Fig.~\ref{fig:abla}, our ablation study evaluates both model architecture and training loss. In Fig.~\ref{fig:abla}(a), the V2V baseline fails to accurately localize the editing region and often modifies the wrong text. MV2V improves region targeting but still suffers from glyph errors. Introducing dual-granularity glyph conditioning corrects the text content, although style inconsistency remains. Finally, equipping the model with VLM enables precise text editing while preserving style consistency. In Fig.~\ref{fig:abla}(b), the flow matching loss alone leads to suboptimal text accuracy and style quality. Adding the OCR loss improves text accuracy but degrades style consistency. The mask reweighting loss enhances style consistency yet still leaves minor character errors. Our GLAS loss combines the strengths of both and achieves the best trade-off between text accuracy and style consistency.

\section{Additional Qualitative Comparison}
\label{appx:comparison}
\begin{figure}[h] % h表示here，即尽量放在当前位置
    \centering
    \includegraphics[width=0.84\textwidth]{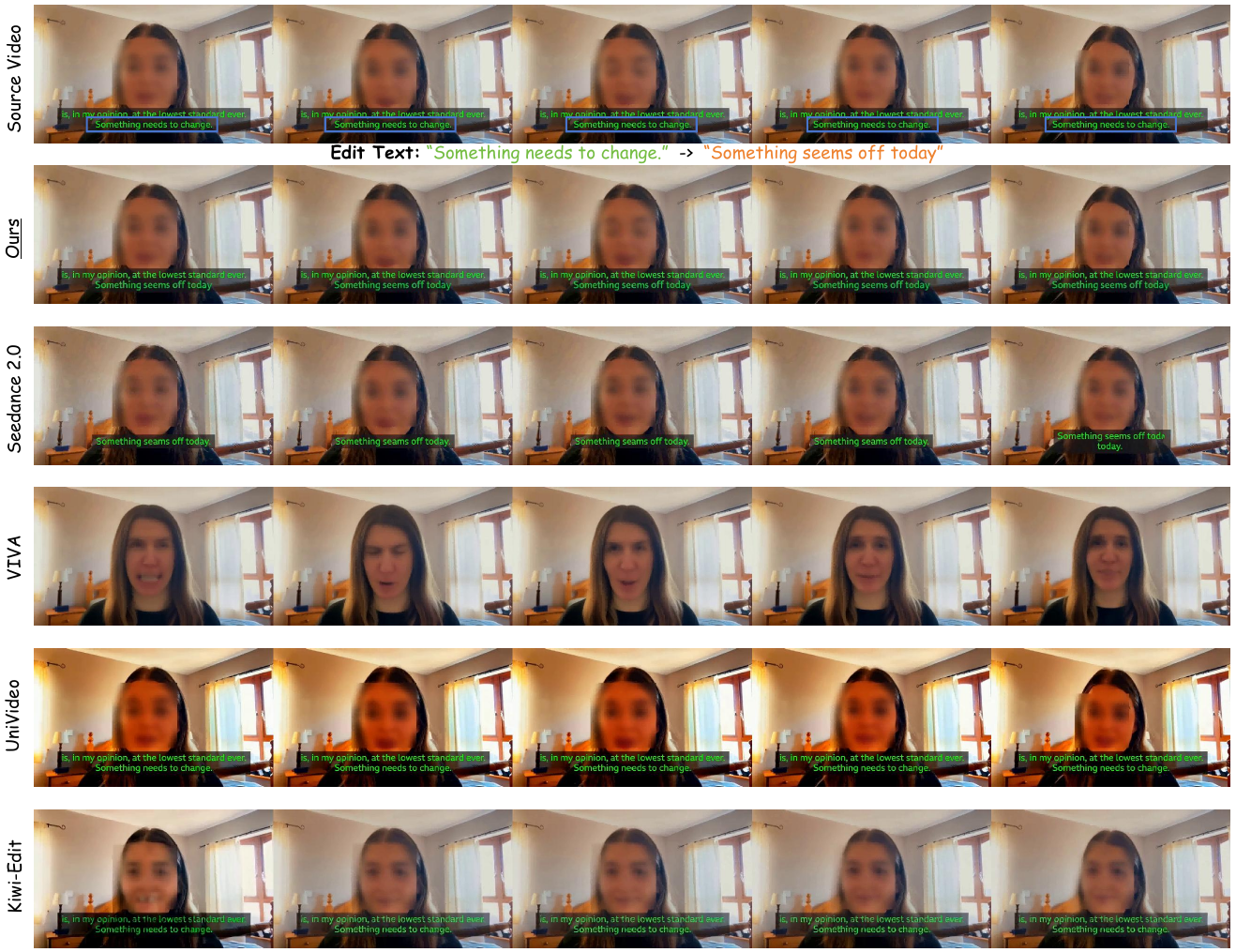} % 图片路径和大小
    % \vspace{-3mm}
    \caption{Visual comparison on real-world scenarios.}
    \vspace{-2mm}
    \label{fig:appendix_comparison_10}
\end{figure}

\begin{figure}[h] % h表示here，即尽量放在当前位置
    \centering
    \includegraphics[width=0.84\textwidth]{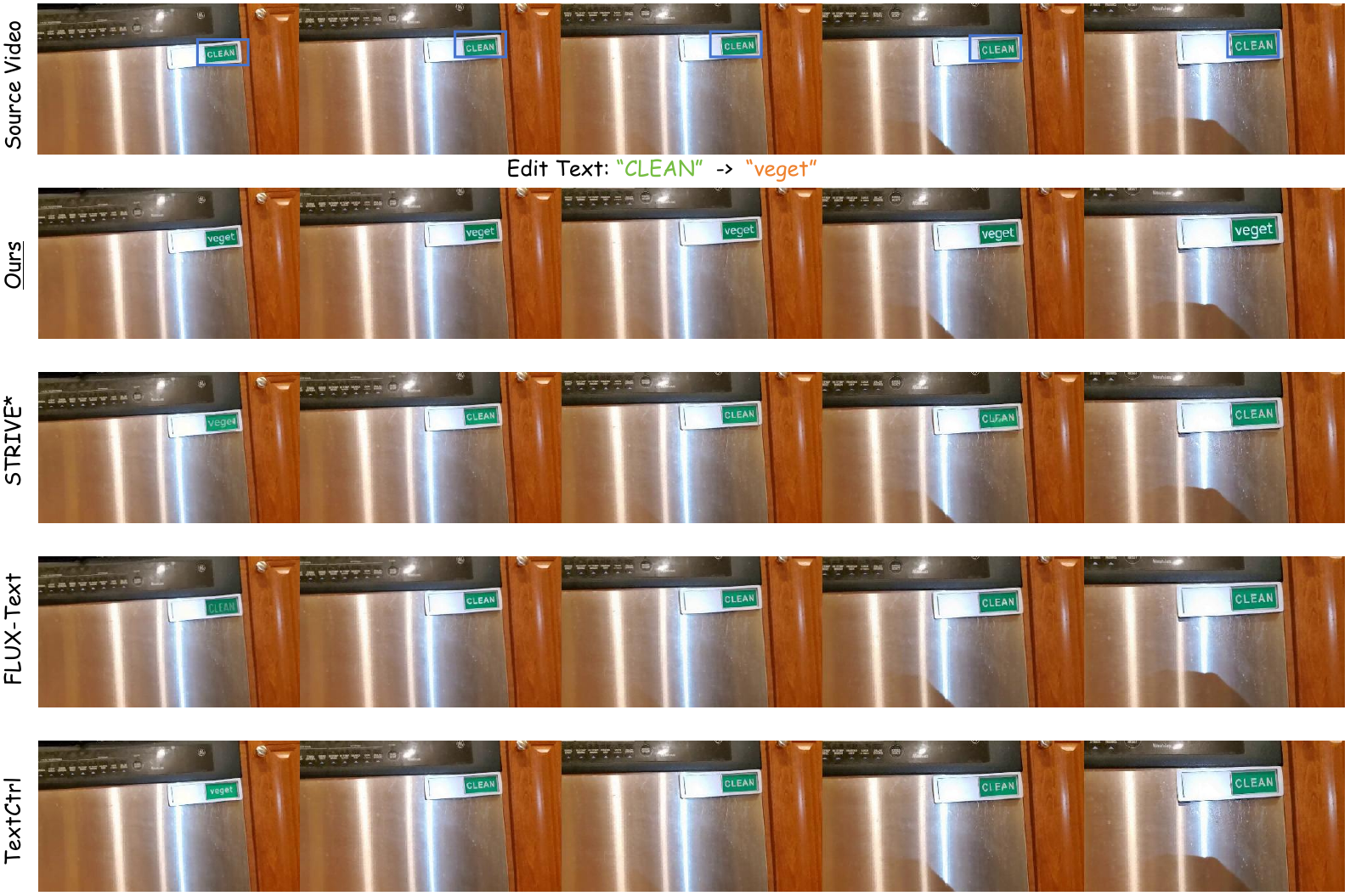} % 图片路径和大小
    % \vspace{-3mm}
    \caption{Visual comparison on scenetext scenarios.}
    \vspace{-2mm}
    \label{fig:scenetext_comparision_1}
\end{figure}

\begin{figure}[h] % h表示here，即尽量放在当前位置
    \centering
    \includegraphics[width=0.84\textwidth]{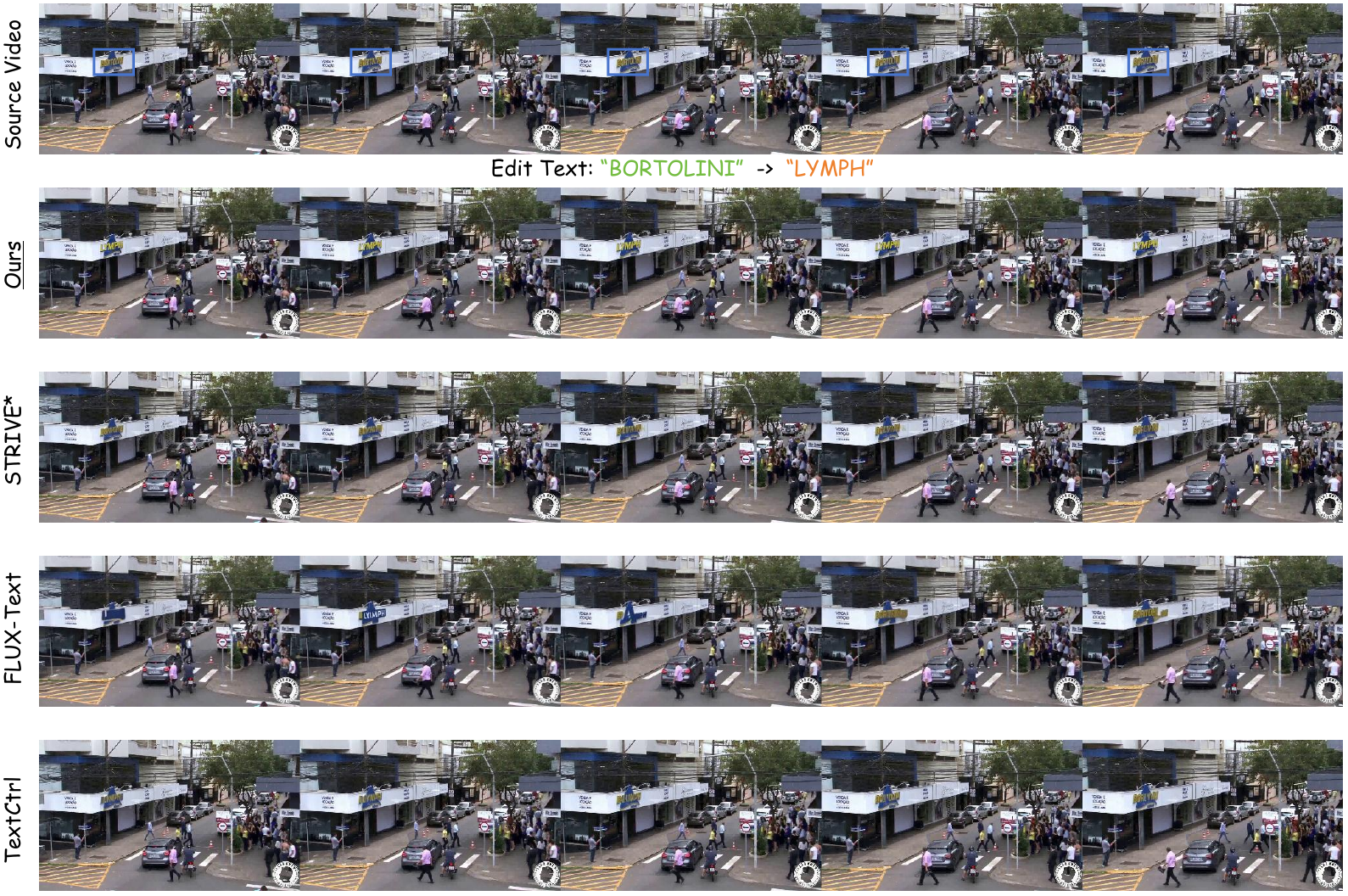} % 图片路径和大小
    % \vspace{-3mm}
    \caption{Visual comparison on scenetext scenarios.}
    \vspace{-2mm}
    \label{fig:scenetext_comparision_2}
\end{figure}

\begin{figure}[h] % h表示here，即尽量放在当前位置
    \centering
    \includegraphics[width=0.84\textwidth]{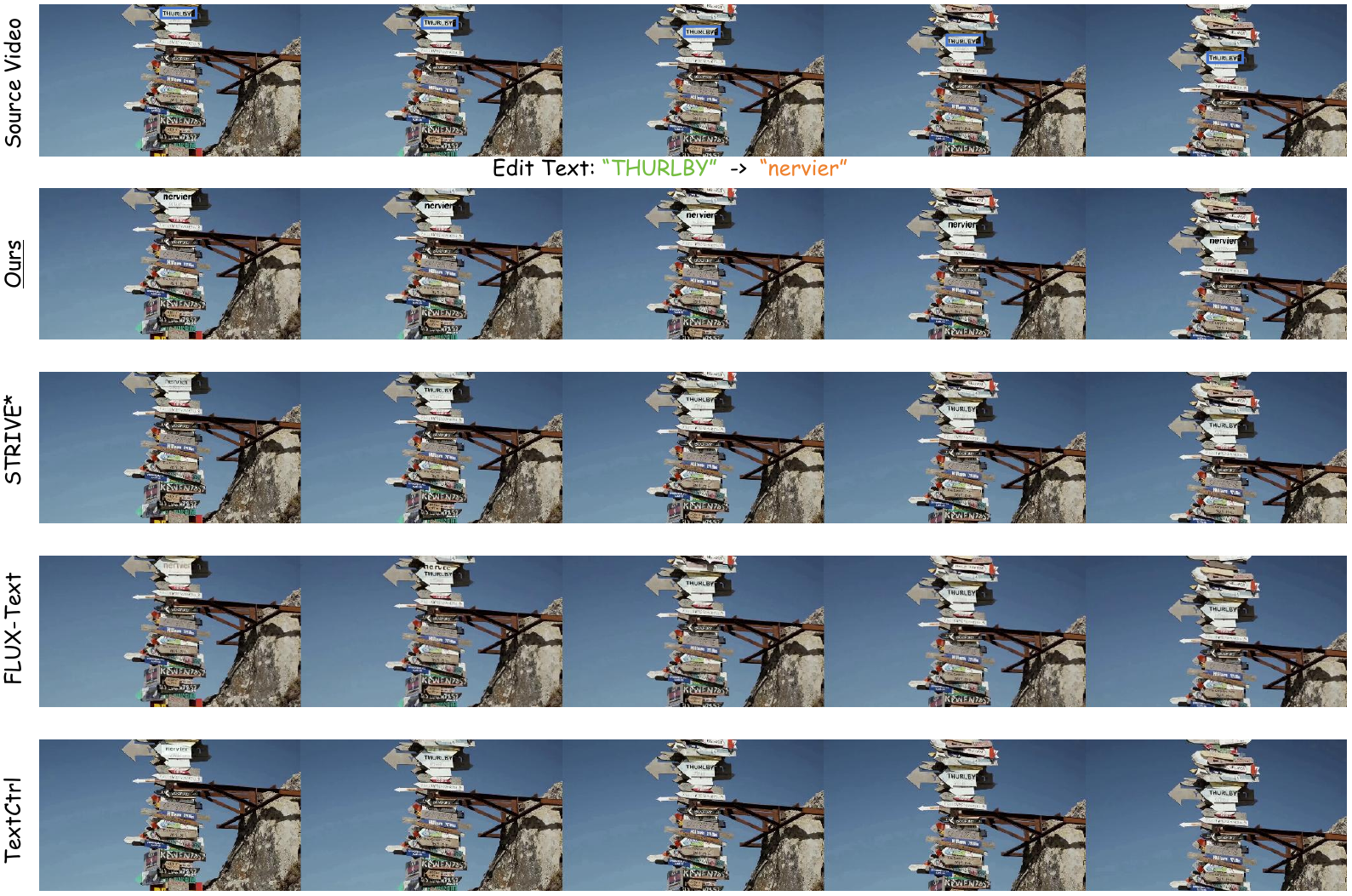} % 图片路径和大小
    % \vspace{-3mm}
    \caption{Visual comparison on scenetext scenarios.}
    \vspace{-2mm}
    \label{fig:scenetext_comparision_3}
\end{figure}

\begin{figure}[h] % h表示here，即尽量放在当前位置
    \centering
    \includegraphics[width=0.84\textwidth]{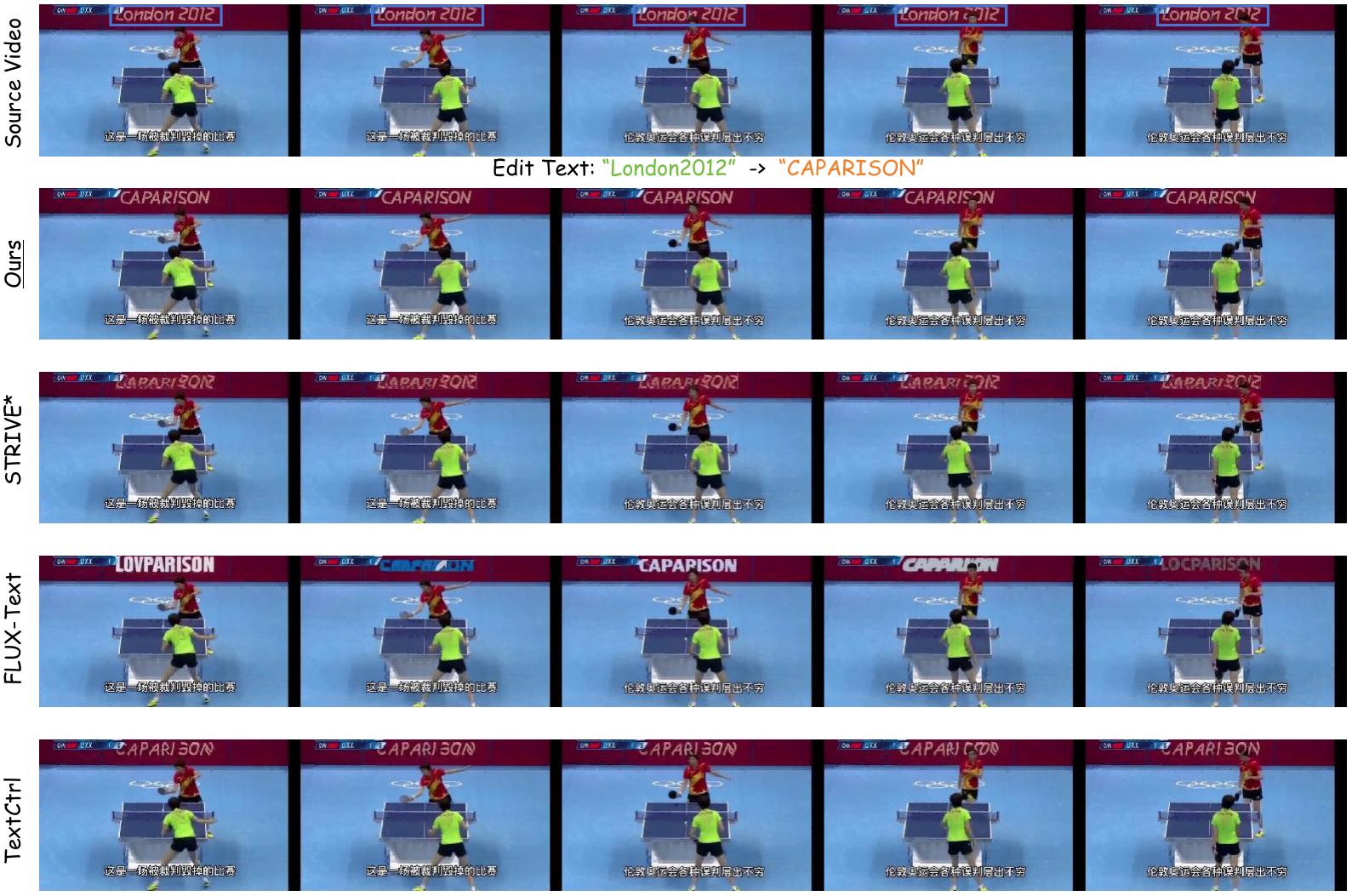} % 图片路径和大小
    % \vspace{-3mm}
    \caption{Visual comparison on scenetext scenarios.}
    \vspace{-2mm}
    \label{fig:scenetext_comparision_4}
\end{figure}

We provide additional qualitative comparisons between our method and other video editing methods on all synthetic, real-world and scenetext scenarios in Figures~\ref{fig:appendix_comparison_10}--\ref{fig:scenetext_comparision_4}. For a better visual experience, please visit our project page \url{https://zengkaiya.github.io/SteerVTE/}.

\section{Broader Impacts}
SteerVTE has the potential to advance research on fine-grained and controllable video editing by enabling precise text modification in videos while preserving stylistic consistency, visual realism, and temporal coherence. Such capabilities may benefit applications including video localization, media production, accessibility, and creative content editing. At the same time, realistic video text editing techniques could be misused to manipulate on-screen information in deceptive ways, potentially contributing to misinformation or fraud. 
In addition, performance may vary across languages, writing systems, fonts, and visual contexts due to biases in training data. 
% We therefore encourage responsible use of this technology, along with continued research on robustness, fairness, and methods for detecting edited media.
To mitigate these concerns, we emphasize that research on video text editing should be accompanied by responsible deployment practices, including usage restrictions, transparency mechanisms, and continued study of detection and attribution methods for edited media. By releasing our findings responsibly, we aim to support progress in controllable video editing while promoting ethical consideration of its societal impact.

\newpage
% \begin{figure}[h] % h表示here，即尽量放在当前位置
%     \centering
%     \includegraphics[width=0.84\textwidth]{figs/appendix_comparison_1.pdf} % 图片路径和大小
%     % \vspace{-3mm}
%     \caption{Visual comparison on synthetic scenarios.}
%     \vspace{-2mm}
%     \label{fig:appendix_comparison_1}
% \end{figure}

% \begin{figure}[h] % h表示here，即尽量放在当前位置
%     \centering
%     \includegraphics[width=0.84\textwidth]{figs/appendix_comparison_2.pdf} % 图片路径和大小
%     % \vspace{-3mm}
%     \caption{Visual comparison on synthetic scenarios.}
%     \vspace{-2mm}
%     \label{fig:appendix_comparison_2}
% \end{figure}

% \begin{figure}[h] % h表示here，即尽量放在当前位置
%     \centering
%     \includegraphics[width=0.84\textwidth]{figs/appendix_comparison_3.pdf} % 图片路径和大小
%     % \vspace{-3mm}
%     \caption{Visual comparison on synthetic scenarios.}
%     \vspace{-2mm}
%     \label{fig:appendix_comparison_3}
% \end{figure}

\begin{figure}[h] % h表示here，即尽量放在当前位置
    \centering
    \includegraphics[width=0.84\textwidth]{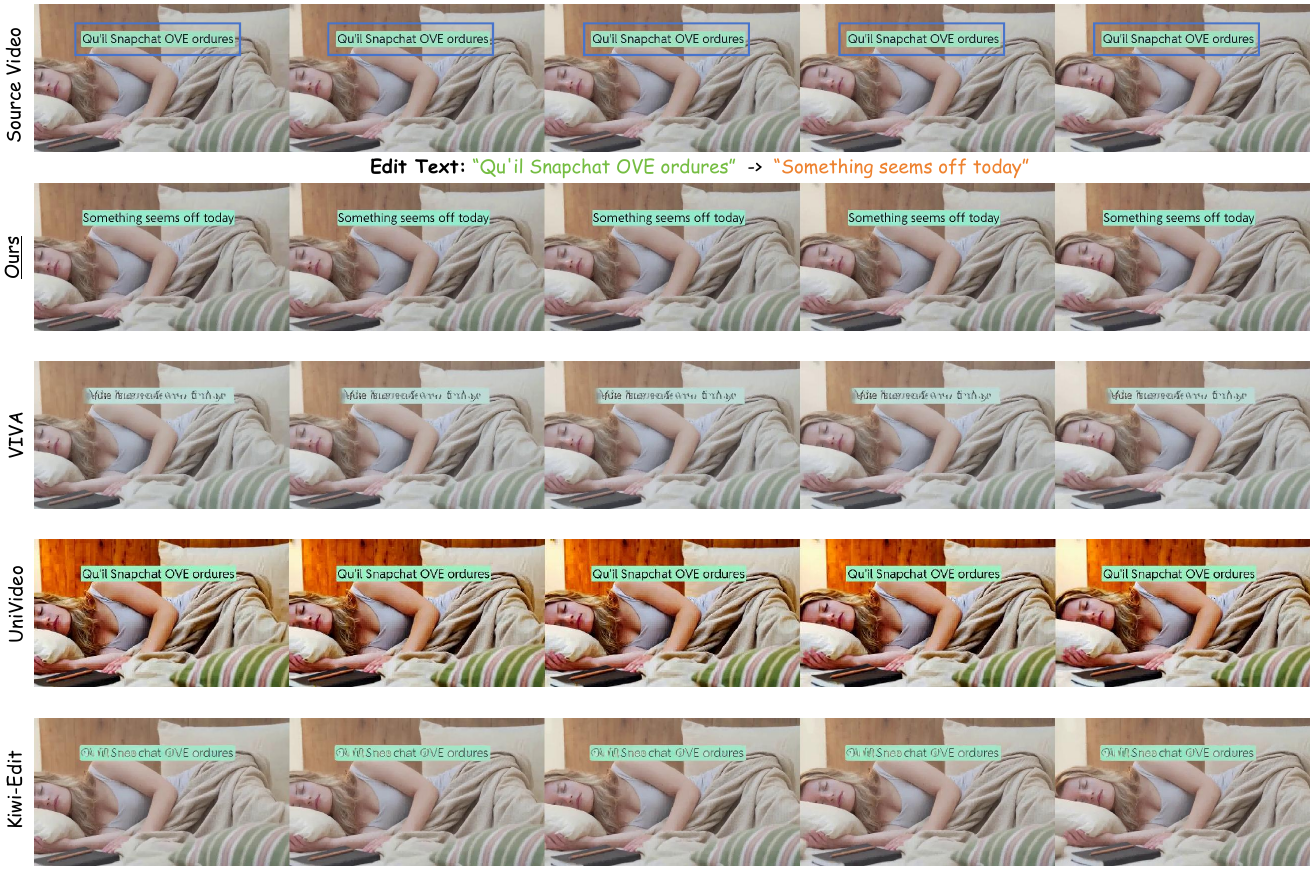} % 图片路径和大小
    % \vspace{-3mm}
    \caption{Visual comparison on synthetic scenarios.}
    \vspace{-2mm}
    \label{fig:appendix_comparison_4}
\end{figure}

% \begin{figure}[h] % h表示here，即尽量放在当前位置
%     \centering
%     \includegraphics[width=0.84\textwidth]{figs/appendix_comparison_5.pdf} % 图片路径和大小
%     % \vspace{-3mm}
%     \caption{Visual comparison on synthetic scenarios.}
%     \vspace{-2mm}
%     \label{fig:appendix_comparison_5}
% \end{figure}

% \begin{figure}[h] % h表示here，即尽量放在当前位置
%     \centering
%     \includegraphics[width=0.84\textwidth]{figs/appendix_comparison_6.pdf} % 图片路径和大小
%     % \vspace{-3mm}
%     \caption{Visual comparison on real-world scenarios.}
%     \vspace{-2mm}
%     \label{fig:appendix_comparison_6}
% \end{figure}

% \begin{figure}[h] % h表示here，即尽量放在当前位置
%     \centering
%     \includegraphics[width=0.84\textwidth]{figs/appendix_comparison_7.pdf} % 图片路径和大小
%     % \vspace{-3mm}
%     \caption{Visual comparison on real-world scenarios.}
%     \vspace{-2mm}
%     \label{fig:appendix_comparison_7}
% \end{figure}

% \begin{figure}[h] % h表示here，即尽量放在当前位置
%     \centering
%     \includegraphics[width=0.84\textwidth]{figs/appendix_comparison_8.pdf} % 图片路径和大小
%     % \vspace{-3mm}
%     \caption{Visual comparison on real-world scenarios.}
%     \vspace{-2mm}
%     \label{fig:appendix_comparison_8}
% \end{figure}

% \begin{figure}[h] % h表示here，即尽量放在当前位置
%     \centering
%     \includegraphics[width=0.84\textwidth]{figs/appendix_comparison_9.pdf} % 图片路径和大小
%     % \vspace{-3mm}
%     \caption{Visual comparison on real-world scenarios.}
%     \vspace{-2mm}
%     \label{fig:appendix_comparison_9}
% \end{figure}

% \begin{figure}[h] % h表示here，即尽量放在当前位置
%     \centering
%     \includegraphics[width=0.84\textwidth]{figs/appendix_dataset_2.pdf} % 图片路径和大小
%     % \vspace{-3mm}
%     \caption{Examples from our stage-1 image training dataset: SynthTE-Easy.}
%     \vspace{-2mm}
%     \label{fig:appendix_dataset_2}
% \end{figure}

% \begin{figure}[h] % h表示here，即尽量放在当前位置
%     \centering
%     \includegraphics[width=0.7\textwidth]{figs/appendix_dataset_4.pdf} % 图片路径和大小
%     % \vspace{-3mm}
%     \caption{Examples from our stage-2 image training dataset: RealTE.}
%     \vspace{-2mm}
%     \label{fig:appendix_dataset_4}
% \end{figure}

% \begin{figure}[h] % h表示here，即尽量放在当前位置
%     \centering
%     \includegraphics[width=0.84\textwidth]{figs/appendix_dataset_5.pdf} % 图片路径和大小
%     % \vspace{-3mm}
%     \caption{Examples from our stage-3 video training dataset: SynthTE-Hard.}
%     \vspace{-2mm}
%     \label{fig:appendix_dataset_5}
% \end{figure}

% \begin{figure}[h] % h表示here，即尽量放在当前位置
%     \centering
%     \includegraphics[width=0.84\textwidth]{figs/appendix_dataset_7.pdf} % 图片路径和大小
%     % \vspace{-3mm}
%     \caption{Examples from our stage-3 video training dataset: SynthTE-Hard.}
%     \vspace{-2mm}
%     \label{fig:appendix_dataset_7}
% \end{figure}

% \section{Technical appendices and supplementary material}
% Technical appendices with additional results, figures, graphs, and proofs may be submitted with the paper submission before the full submission deadline (see above). You can upload a ZIP file for videos or code, but do not upload a separate PDF file for the appendix. There is no page limit for the technical appendices. 

% Note: Think of the appendix as ``optional reading'' for reviewers. The paper must be able to stand alone without the appendix; for example, adding critical experiments that support the main claims to an appendix is inappropriate. 

%%%%%%%%%%%%%%%%%%%%%%%%%%%%%%%%%%%%%%%%%%%%%%%%%%%%%%%%%%%%
% \clearpage
% \newpage
% \input{checklist.tex}

\end{document}